\documentclass[10pt,twocolumn,letterpaper]{article}

\usepackage{cvpr}              

\usepackage{graphicx}
\usepackage{amsmath}
\usepackage{amssymb}
\usepackage{booktabs}
\usepackage{subcaption,amsfonts,dcolumn}
\usepackage[pagebackref,breaklinks,colorlinks]{hyperref}
\usepackage{wrapfig,lipsum,booktabs}
\usepackage{graphicx}
\usepackage{tikz}
\usepackage{comment}
\usepackage{amsmath,amssymb}
\usepackage{color}
\usepackage{epsfig}
\usepackage{multirow}
\usepackage{enumitem}
\usepackage{wrapfig}
\usepackage{textcomp,booktabs}
\usepackage{xcolor}
\usepackage[tablename=Table]{caption} 
\usepackage{dsfont}
\usepackage{makecell}
\usepackage{array}
\usepackage{algorithm}
\usepackage{hyperref}
\usepackage{algorithm}
\usepackage{algpseudocode}
\usepackage[accsupp]{axessibility}
\newcolumntype{x}[1]{>{\centering\arraybackslash\hspace{0pt}}p{#1}}

\definecolor{green}{rgb}{0.0, 0.5, 0.0}

\newcommand{\myparagraph}[1]{\vspace{1pt}\noindent{\bf{#1}}}

\begin{document}

\title{ScaleDet: A Scalable Multi-Dataset Object Detector}

\author{
Yanbei Chen, \hspace{3pt} Manchen Wang, \hspace{3pt} Abhay Mittal, \hspace{3pt} Zhenlin Xu, \\ Paolo Favaro, \hspace{3pt} Joseph Tighe, \hspace{3pt} Davide Modolo \\
AWS AI Labs
}

\maketitle

\begin{abstract}
Multi-dataset training provides a viable solution for exploiting heterogeneous large-scale datasets without extra annotation cost. In this work, we propose a scalable multi-dataset detector (ScaleDet) that can scale up its generalization across datasets when increasing the number of training datasets. Unlike existing multi-dataset learners that mostly rely on manual relabelling efforts or sophisticated optimizations to unify labels across datasets, we introduce a simple yet scalable formulation to derive a unified semantic label space for multi-dataset training. ScaleDet is trained by visual-textual alignment to learn the label assignment with label semantic similarities across datasets. Once trained, ScaleDet can generalize well on any given upstream and downstream datasets with seen and unseen classes. We conduct extensive experiments using LVIS, COCO, Objects365, OpenImages as upstream datasets, and 13 datasets from Object Detection in the Wild (ODinW) as downstream datasets. Our results show that ScaleDet achieves compelling strong model performance with an mAP of 50.7 on LVIS, 58.8 on COCO, 46.8 on Objects365, 76.2 on OpenImages, and 71.8 on ODinW, surpassing state-of-the-art detectors with the same backbone. 
\end{abstract}

\vspace{-1em}
\section{Introduction}
\label{sec:intro}

Major advances in computer vision have been driven by large-scale datasets, such as ImageNet \cite{deng2009imagenet} and OpenImages \cite{OpenImages2} for image classification, or Kinetics \cite{carreira2017quo} and ActivityNet \cite{caba2015activitynet} for video recognition. Large-scale datasets are crucial for training recognition models that generalize well. However, the collection of massive annotated datasets is costly and time-consuming. This is especially prominent in detection and segmentation tasks that require detailed annotations at the bounding box or pixel level. To exploit more training data without extra annotation cost, recent works unify multiple datasets to learn from more visual categories and more diverse visual domains for detection \cite{wang2019towards,zhao2020object,xu2020universal,zhou2022simple} and segmentation \cite{lambert2020mseg,uijlings2022missing}.

\begin{figure}[!t]
  \centering
\includegraphics[width=0.85\linewidth]{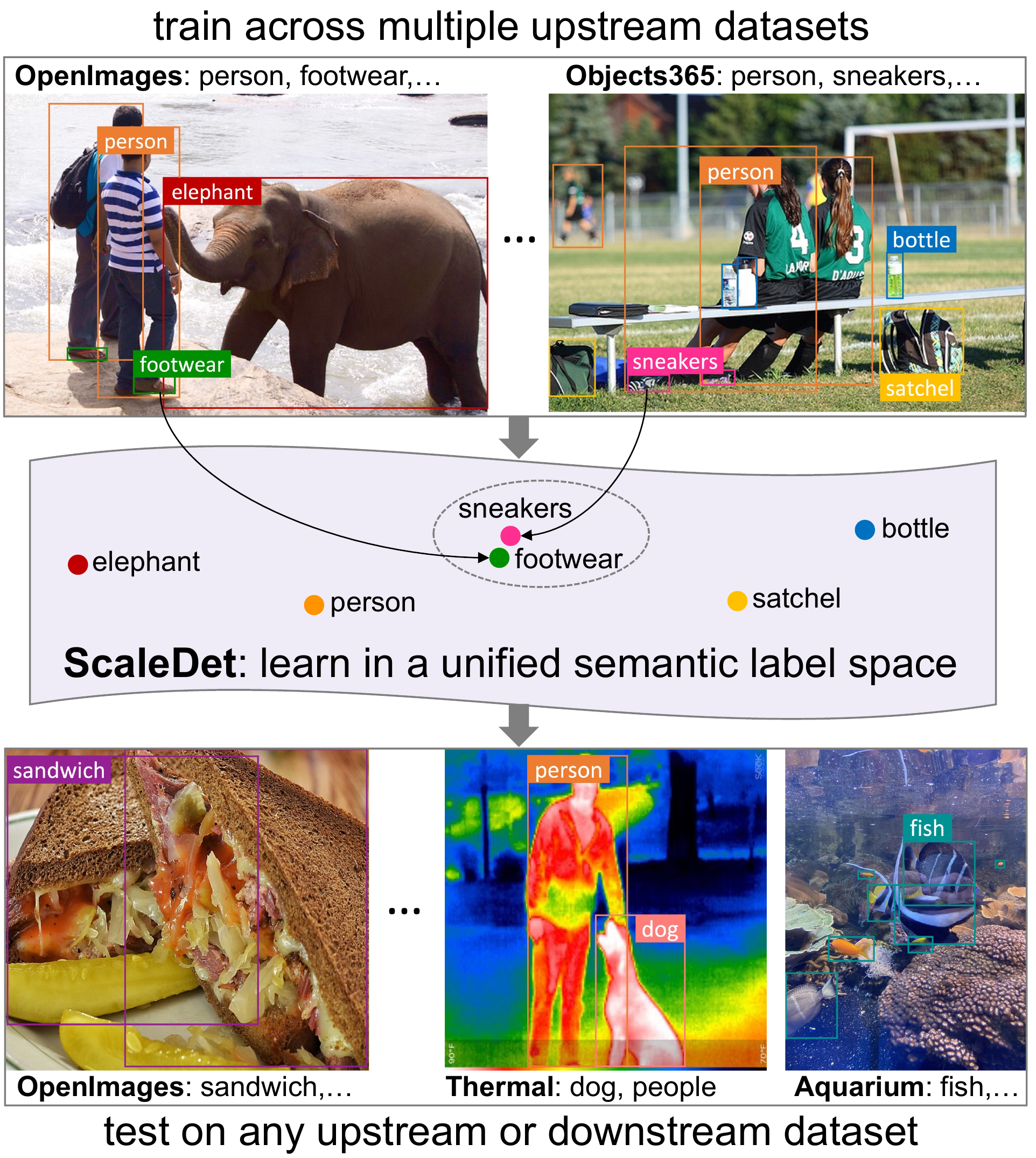}
   \vskip -0.8em
   \caption{
Our scalable multi-dataset detector (ScaleDet) learns across datasets in a unified semantic label space by visual-textual alignment with label semantic similarities. At test time, ScaleDet can generalize on any given upstream and downstream dataset. 
   }
   \label{fig:teaser}
   \vskip -1em
\end{figure}

To train an object detector across multiple datasets, we need to tackle several challenges.
{\it First}, multi-dataset training requires unifying the heterogeneous label spaces across datasets, as label definitions are dataset-specific. The labels from two datasets may indicate the same or similar objects. {For example,} ``{footwear}'' and ``{sneakers}'' are two different labels in OpenImages \cite{kuznetsova2020open} and Objects365 \cite{shao2019objects365}, but refer to the same type of objects (see Figure~\ref{fig:teaser}).
{\it Second}, the training setups may be inconsistent among datasets, as different data sampling strategies and learning schedules are often required for datasets of different sizes.
{\it Third}, a multi-dataset model should perform better than single-dataset models on individual datasets. This is challenging due to the heterogeneous label spaces, the domain discrepancy across datasets, and the risk of overfitting to the larger datasets.

To resolve the above challenges, existing work resorts to manually relabelings class labels \cite{lambert2020mseg}, or training multiple dataset-specific classifiers with constraints to relate labels across datasets \cite{zhou2022simple}. However, these methods lack scalability. The manual relabeling effort and the model complexity of training multiple classifiers grow rapidly as the number of datasets increases. 
We overcome this limitation with {\bf ScaleDet}: a scalable multi-dataset detector 
(Figure \ref{fig:teaser}). 
We propose two innovations: {\em a scalable formulation} to unify multiple label spaces, and {\em a novel loss formulation} to learn hard label and soft label assignments across datasets. 
While hard label assignment serves to disambiguate class labels in probability space, soft label assignment works as a regularizer to relate class labels in semantic similarity space. 
Unlike existing multi-dataset methods \cite{wang2019towards,zhao2020object,xu2020universal,zhou2022simple,lambert2020mseg,uijlings2022missing} that mostly generalize on seen datasets or seen classes, 
our method exploits vision-language learning to attain good generalization on both upstream and downstream datasets, where the downstream datasets can contain unseen classes and new domains.
Our contributions are:
\vspace{-0.5em}
\begin{itemize}[noitemsep,leftmargin=6pt,font=\bfseries]
\item We propose a novel scalable multi-dataset training recipe for object detection. Our method utilizes text embeddings to unify and relate labels with semantic similarities across datasets, and trains a single classifier via visual-textual alignment to learn hard label and soft label assignments. 
\item We conduct extensive experiments to demonstrate the compelling scalability and generalizability of ScaleDet in multi-dataset training. We show that ScaleDet can boost its performance as we increase the number of training datasets: LVIS \cite{gupta2019lvis}, COCO \cite{lin2014microsoft}, Objects365 \cite{shao2019objects365} and OpenImages \cite{kuznetsova2020open} (Sec \ref{sec:analysis}). Furthermore, we show that ScaleDet achieves state-of-the-art performance on multiple benchmarks when compared to recent advanced detectors, e.g., Detic~\cite{zhou2022detecting}, UniDet~\cite{zhou2022simple} (Sec \ref{sec:multisota}, Sec \ref{sec:sota}). 
\item We evaluate the transferablity of ScaleDet on the challenging ``Object Detection in the Wild'' benchmark (which contains 13 datasets) \cite{li2022grounded} to demontrate its competitive generalizability on downstream datasets (Sec \ref{sec:odinw}). 
\end{itemize}

\section{Related Work}
\label{sec:related}

\myparagraph{Multi-dataset training} aims to leverage multiple datasets to train more generalizable visual recognition models, especially for tasks that require more expensive annotations, such as detection \cite{wang2019towards,zhao2020object,xu2020universal,zhou2022simple} and segmentation \cite{lambert2020mseg,uijlings2022missing}. 
Existing methods for multi-dataset training can be categorized into two groups. The first group introduces special network components to adapt feature representations tailored to datasets \cite{wang2019towards,xu2020universal}. For example, a domain attention module is designed to learn an adaptive multi-domain detector, which assigns different network activations to different domains \cite{wang2019towards}. The second group introduces new formulations to combine heterogeneous label spaces over multiple datasets \cite{lambert2020mseg,zhao2020object,zhou2022simple,uijlings2022missing}. For instance, the MSeg dataset \cite{lambert2020mseg} is created upon multiple semantic segmentation datasets by relabeling with Amazon Mechanical Turk based on a manually defined taxonomy of class labels. To avoid manual relabeling, a pseudo labeling strategy \cite{zhao2020object} is used to generate pseudo labels across datasets based on predictions from dataset-specific detectors. Recent works use set theory with manually defined sets \cite{uijlings2022missing} or employ combinatorial optimization \cite{zhou2022simple} to learn the label relations across datasets. 
Our approach is more related to the second group, but it offers a more scalable and generalizable training recipe that does not require training multiple dataset-specific classifiers \cite{zhao2020object,zhou2022simple}, nor manually designed rules to relate class labels \cite{lambert2020mseg,uijlings2022missing}. 
{Furthermore, different from existing multi-dataset detectors \cite{wang2019towards,zhao2020object,xu2020universal,zhou2022simple}, ScaleDet is also capable of generalizing to datasets that contain unseen classes. }

\myparagraph{Vision-language models} (VLMs) employ vision-and-language learning to solve visual recognition problems. 
By bridging large-scale vision and language data, vision-language models (VLMs) such as VirTex \cite{desai2021virtex}, CLIP \cite{radford2021learning}, ALIGN \cite{jia2021scaling}, FILIP \cite{yao2021filip}, UniCL \cite{yang2022unified} and LiT \cite{zhai2022lit}, have shown remarkable performance on learning transferable visual representations that generalize well to downstream tasks. 
More recently, VLMs have been explored in segmentation \cite{xu2022groupvit,ding2022open} and detection \cite{kamath2021mdetr,gu2021open,zhong2022regionclip,li2022grounded,zhang2022glipv2,cai2022x,zhou2022detecting}. In detection, most of the VLMs utilize auxiliary semantic-rich vision-language datasets as pre-training data to build models capable of solving multiple tasks including open-vocabulary detection \cite{zareian2021open}, as represented by MDETR \cite{kamath2021mdetr}, RegionCLIP \cite{zhong2022regionclip}, GLIP \cite{li2022grounded,zhang2022glipv2}, X-DETR \cite{cai2022x}, and Detic \cite{zhou2022detecting}. For instance, MDETR uses various vision-and-language datasets (e.g., Flickr30k \cite{plummer2015flickr30k}, Visual genome \cite{krishna2017visual}) to train a VLM that allows instance-wise detection guided by text query. GLIP \cite{li2022grounded} uses 27M grounding data to build a unified VLM for detection and grounding. X-DETR \cite{cai2022x} trains a VLM on grounding and image-caption datasets to solve multiple instance-wise vision-language tasks in one model. Detic \cite{zhou2022detecting} uses detection and large-scale classification datasets to train a large-vocabulary detector. 
We exploit a pre-trained text encoder similar to other VLMs, e.g., RegionCLIP uses CLIP, while MDETR, X-DETER use RoBERTa. However, our model is trained directly on detection datasets, while these other VLMs use image captioning or grounding datasets to train jointly with detection datasets.

\section{ScaleDet: A Scalable Multi-Dataset Detector}
\label{sec:method}

\begin{figure*}[!t]
  \centering
   \includegraphics[width=0.96\linewidth]{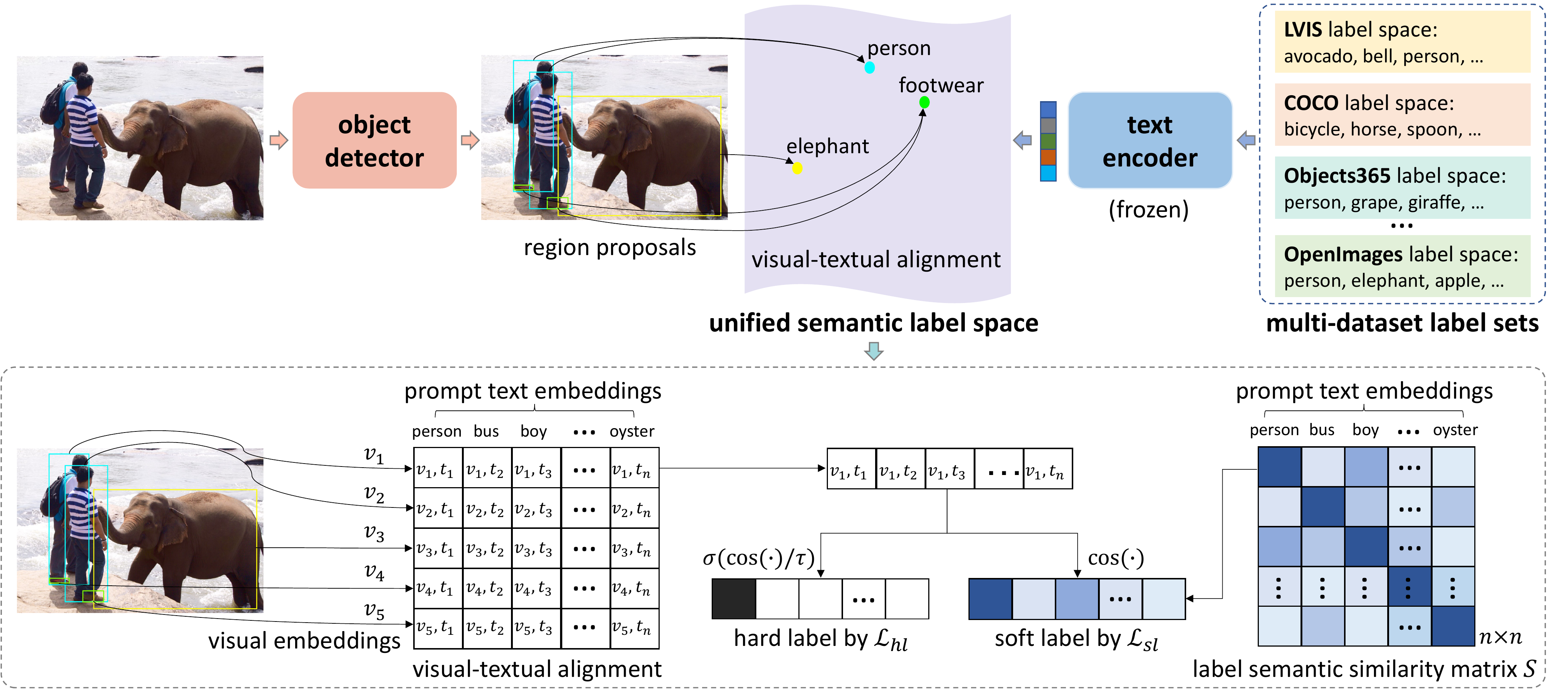}
   \vskip -0.5em
   \caption{{\bf Approach overview.} Our scalable multi-dataset detector (ScaleDet) uses text prompts to define labels, and learns across datasets by visual-textual alignment in the unified label space. Each visual feature of an object region proposal is assigned to one corresponding label {via a hard label assignment (see the loss $\mathcal{L}_{hl}$ in Eq. \eqref{eq:hard}) and related to semantically-similar labels via a soft label assignment (see the loss $\mathcal{L}_{sl}$ in Eq. \eqref{eq:soft}).} At test time, ScaleDet can perform object detection on any given label space of a seen or unseen detection dataset.}
   \label{fig:model}
   \vskip -1em
\end{figure*}

{Our goal is to train object detectors that generalize better as we increase the number of datasets used for training.}
A critical challenge of multi-dataset training is unifying the heterogeneous label spaces across datasets by relating their semantic concepts. 
To tackle this challenge,
we propose a simple yet effective recipe to train a scalable multi-dataset detector (ScaleDet, Figure \ref{fig:model}). 
ScaleDet learns across multiple datasets by unifying different label sets to form a unified semantic label space (Figure \ref{fig:model} top), and 
{is trained via hard label and soft label assignments} 
for visual-textual alignment (Figure~\ref{fig:model} bottom). 
In Sec~\ref{sec:model} and Sec~\ref{sec:training} we present the details of these two contributions, but first we discuss the preliminaries and problem formulation in Sec~\ref{sec:pre}. 

\subsection{Preliminaries and problem formulation} 
\label{sec:pre}

\myparagraph{Standard object detection.}
A typical object detector aims to predict the bounding box location $b_i \in \mathbf{R}^4$ and class label $c_i \in \mathbb{R}^n$ for any object that belongs to a predefined set of $n$ classes. 
Given an image $I$, the image encoder (e.g., CNN or Transformer) of the detector extracts the box features and visual features, which are fed to a box regressor $B$ and a visual classifier $C$. The detector is trained by {minimizing} a bounding box regression loss $\mathcal{L}_{bbox}$ and a classification loss $\mathcal{L}_{cls}$ to learn the predictions of bounding boxes and class labels given the box features and visual features, i.e., 
\begin{equation}
\begin{aligned}
\mathcal{L}_{Det} = \mathcal{L}_{bbox} + \mathcal{L}_{cls}.
\end{aligned}
\label{eq:basic}
\end{equation}
Existing object detectors generally adopt a one-stage \cite{tian2019fcos,zhou2019objects} or two-stage \cite{ren2015faster,he2017mask} framework, which may contain additional loss terms in Eq.~(\ref{eq:basic}). For example, one-stage detectors use a regression loss to regress towards properties of object locations, such as centerness \cite{tian2019fcos}. Two-stage detectors \cite{ren2015faster,he2017mask} instead use a region proposal network with its own dedicated loss function that predicts the objectness of each box. 
In this work, we focus on reformulating the classification loss $\mathcal{L}_{cls}$ in Eq.~\eqref{eq:basic} to solve the problem of multi-dataset training, 
built upon a two-stage detector \cite{zhou2021probabilistic}.

\myparagraph{Multi-dataset object detection.}  
Given a set of $K$ datasets $\{\mathcal{D}_1, \mathcal{D}_2, ..., \mathcal{D}_K\}$ with their label spaces $\{L_1, L_2, ..., L_K\}$, we aim at training a scalable multi-dataset detector that generalizes well on upstream and downstream detection datasets. 
In contrast to prior multi-dataset learners \cite{lambert2020mseg,zhou2022simple,uijlings2022missing}, 
which relate or merge similar labels across datasets to joint labels, we propose a simple yet scalable formulation to unify labels without merging any labels explicitly. 
We exploit informative text embeddings from powerful pre-trained visual-language models to define and relate labels over the non-identical label spaces $\{L_1, L_2, ..., L_K\}$. We present our method in the following sections. 

\subsection{Scalable unification of multi-dataset label space}
\label{sec:model}

As Figure~\ref{fig:model}~{(top)} shows, 
during training, a mini-batch of images is randomly sampled from multiple training sets and fed to the detector to obtain the visual features $\{v_1, v_2, ..., v_j\}$ 
of the region proposals for each image, 
where $v_i \in \mathbf{R}^D$ is a $D$-dimensional vector. Each visual feature $v_i$ is matched to a set of text embeddings $\{t_1, t_2, ..., t_n\}$ by label assignment. Below, we detail how we define semantic labels with text prompts, and unify label spaces with label semantic similarities for multi-dataset training. 

\myparagraph{Define labels with text prompts.} We represent each class label $l_i$ with its text prompts, e.g., the label {\it ``person''} can be denoted by a text prompt {\it ``a photo of a person''}. We extract the prompt text embeddings from the text encoder of a pre-trained visual-language model (e.g., CLIP \cite{radford2021learning} or OpenCLIP \cite{ilharco_gabriel_2021_5143773}), and average embeddings of a set of predefined text prompts (known as {\em prompt engineering} \cite{radford2021learning}) to represent the label $l_i$ as a semantic text embedding $t_i$.

\myparagraph{Unify label spaces by concatenation.} Given the text embeddings of class labels from all datasets, a crucial problem of multi-dataset training is to unify the non-identical label spaces $\{L_1, L_2, ..., L_K\}$, which can be solved by relating and merging similar labels into a unified label set \cite{lambert2020mseg}. However, without careful manual inspection, this incurs the risk of propagating errors in model training due to the ambiguities in label definitions, e.g., labels ``{\em boy}'' and ``{\em girl}'' are similar but should not be merged. Thus, instead of merging labels across datasets, we unify different label spaces by concatenation: 
\begin{equation}
\begin{aligned}
L {=} L_1{\coprod}{...}{\coprod} L_K 
{=} \{l_{1,1}, l_{1,2},..., l_{K,1}, l_{K,2},... \},
\end{aligned}
\label{eq:concat}
\end{equation} 
where $\coprod$ denotes the coproduct (i.e., the disjoint union of two label sets); $l_{k,i}$ is label $i$ from dataset $k$ (we omit $k$ below). Besides its simplicity, our formulation of this unified semantic label space $L$ maximally preserves the semantics of all labels, 
thus providing richer vocabulary for training. 

\myparagraph{Relate labels by semantic similarities.} 
{As we use text embeddings to represent class labels, we can relate the labels that share similar semantics, in our unified label space. For example, label {\it ``person''} in LVIS \cite{gupta2019lvis} should be related to labels {\it ``person''} and {\it ``boy''} in OpenImages \cite{kuznetsova2020open}, as they are semantically similar.}
To uncover the label relations across datasets, we compute the semantic similarities using the prompt text embeddings. For a given class label $l_i$, its semantic similarities with respect to all the labels are derived with cosine similarities and normalized between 0 and 1:
\begin{equation}
\begin{aligned}
& \text{sim}(l_i,l_j)=\frac{\text{cos}(t_i,t_j)-\alpha_i}{\beta_i-\alpha_i}, \\
\text{where} \
& \alpha_i{=}\text{min}\{ \text{cos}(t_i,t_j) \}_{j=1}^{n}, \\
& \beta_i{=}\text{max}\{ \text{cos}(t_i,t_j) \}_{j=1}^{n}{=}\text{cos}(t_i,t_i){=}1,
\end{aligned}
\label{eq:sim}
\end{equation} 
where $\text{sim}(l_i,l_j)$ is the semantic similarity between the text embeddings $t_i,t_j$ of two labels $l_i,l_j$ and it encodes their semantic relation, e.g., the labels {\it ``person''} and {\it ``boy''} are strongly related with a high similarity score, while {\it ``person''} and {\it ``avocado''} are weakly related with low similarity. 

The label semantic similarity matrix $S$ that encodes label relations among all $n$ class labels can be written as: 
\begin{equation}
\begin{aligned}
S =\left[ \begin{array}{rrr}
1 & \cdots & \text{sim}(l_1,l_n) \\
\vdots & \ddots & \vdots \\
\text{sim}(l_n,l_1)& \cdots & 1 \\
\end{array}\right] 
= \left[ \begin{array}{rrr}
\mathbf{s}_1\\
\vdots \\
\mathbf{s}_n \\
\end{array}\right],
\end{aligned}
\label{eq:matrix}
\end{equation}
where $S$ is an $n\times n$ matrix, and each row vector $\mathbf{s}_i$ encodes the semantic relations of label $l_i$ with respect to all $n$ class labels. With these label semantic similarities, we can introduce explicit {constraints that allow the detector to learn} on the unified semantic label space (Eq.~\eqref{eq:concat}) with encoded label semantic similarities (Eq.~\eqref{eq:matrix}).
Importantly, our formulas (Eq.~\eqref{eq:concat}, Eq.~\eqref{eq:matrix}) are computed offline, 
which does not add any computational cost for training and inference, nor requires a model reformulation when scaling up the number of training datasets.

\subsection{Training with visual-language alignment}
\label{sec:training}

For training on the unified semantic label space $\{l_1, l_2, ..., l_n \}$, we align visual features with text embeddings $\{t_1, t_2, ..., t_n\}$ {via the hard label and soft label assignments,} as shown in Figure~\ref{fig:model} (bottom) and detailed below.

\myparagraph{Visual-language similarities.}
Given the visual feature $v_i$ of an object region proposal, we first compute the cosine similarities between $v_i$ and all text embeddings $\{t_1, t_2, ..., t_n\}$: 
\begin{equation}
\begin{aligned}
\mathbf{c}_i = [ \text{cos}(v_i,t_1), \text{cos}(v_i,t_2), ..., \text{cos}(v_i,t_n)].
\end{aligned}
\label{eq:cos}
\end{equation}
With these similarity scores, we can align the visual feature $v_i$ to the text embeddings with the following loss terms. 

\myparagraph{Hard label assignment.} 
Each visual feature $v_i$ has its groundtruth label $l_i$, and thus can be matched to the text embedding $t_i$ of $l_i$ by hard label assignment as follows, 
\begin{equation}
\begin{aligned}
\mathcal{L}_{hl} = 
\text{BCE}(\sigma_{sg}(\mathbf{c}_i/\tau), l_i),
\end{aligned}
\label{eq:hard}
\end{equation}
where $\text{BCE}(\cdot)$ is the binary cross-entropy loss, $\sigma_{sg}(\cdot)$ is the sigmoid activation, $\tau$ is a temperature hyperparameter. 
Eq.~\eqref{eq:hard} ensures the visual feature $v_i$ is aligned with the text embedding $t_i$. However, it does not explicitly learn the label relations across datasets. We introduce the soft label assignment to learn the semantic label relations. 

\myparagraph{Soft label assignment.} As each label is {semantically} related to all class labels {through the} semantic similarities {computed} in Eq.~\eqref{eq:matrix}, 
{a visual feature can also be related to all text embeddings by using the semantic similarity scores.} 
For this aim, we introduce the soft label assignment on the visual feature $v_i$: 
\begin{equation}
\begin{aligned}
\mathcal{L}_{sl} = \text{MSE}(\mathbf{c}_i, \mathbf{s}_i),
\end{aligned}
\label{eq:soft}
\end{equation}
where $\text{MSE}(\cdot)$ is mean square error. $\mathbf{s}_i$ denotes the semantic similarities between the label $l_i$ and all $n$ class labels -- the $i_{th}$ row of the label semantic similarity matrix $S$ in Eq.~\eqref{eq:sim}.

\noindent
{\bf \em Remark.} While hard label assignment (Eq.~\eqref{eq:hard}) is imposed in probability space to disambiguate different class labels, soft label assignment (Eq.~\eqref{eq:soft}) is imposed in semantic similarity space to assign each visual feature to the text embeddings with different semantic similarities, thus acting as a regularizer to relate similar class labels across datasets. 

\myparagraph{Training with semantic label supervision.} 
With the hard label and soft label assignments, we train the detector to classify different region proposals by aligning visual features to text embeddings in the unified semantic label space. That is, the classification loss $\mathcal{L}_{cls}$ in Eq.~\eqref{eq:basic} is replaced by 
\begin{equation}
\begin{aligned}
\mathcal{L}_{lang} = \mathcal{L}_{hl} + \lambda \mathcal{L}_{sl}, 
\end{aligned}
\label{eq:lang}
\end{equation}
where $\lambda$ is a hyperparameter to balance the two loss terms. Eq. \eqref{eq:lang} uses language supervision to map images to texts, thus enabling zero-shot detection of unseen labels.

\myparagraph{Overall objective.} As we focus on reformulating the classification loss of our detector for multi-dataset training, we do not change the loss $\mathcal{L}_{bbox}$ in Eq.~\eqref{eq:basic} and our overall objective for training ScaleDet is:
\begin{equation}
\begin{aligned}
\mathcal{L}_{ScaleDet} =  \mathcal{L}_{bbox} + \mathcal{L}_{lang}.
\end{aligned}
\label{eq:ScaleDet}
\end{equation} 
Once trained with $\mathcal{L}_{ScaleDet}$, ScaleDet can be deployed on any upstream or downstream datasets that contain seen or unseen classes. By replacing the unified label space $L$ in Eq.~\eqref{eq:concat} with the label space of any given test dataset, ScaleDet can compute the label assignment based on the visual-language similarities derived in Eq.~\eqref{eq:sim}. When the test dataset contains unseen classes, the evaluation setting is known as zero-shot detection \cite{bansal2018zero} or open-vocabulary object detection \cite{zareian2021open}. 
To test on any given dataset, ScaleDet can either be directly evaluated or fine-tuned before evaluation. 

\section{Experiments}
\label{sec:exp}

We detail our experiment setups in Sec~\ref{sec:setup}. We analyze ScaleDet when training with an increasing number of datasets in Sec~\ref{sec:analysis}, and compare it to the state-of-the-art (SOTA) on standard benchmarks in Sec~\ref{sec:multisota} and Sec~\ref{sec:sota}. We evaluate the transferability of ScaleDet on downstream datasets in Sec~\ref{sec:odinw}, and conduct ablation study in Sec~\ref{sec:ablation}. 
We also provide qualitative visual results in {\em Supplementary}.

\subsection{Experiment setups}
\label{sec:setup}

\myparagraph{Upstream datasets.} For our multi-dataset training, we adopt the following four popular detection datasets as uptream datasets: (1) COCO \cite{lin2014microsoft} (C) contains 80 common object categories; (2) LVIS \cite{gupta2019lvis} (L) has a large number of 1203 object categories with a challenging long tail distribution; (3) Objects365 \cite{shao2019objects365} (O365) has 365 object categories; (4) OpenImages detection (OID) \cite{kuznetsova2020open} has 601 object categories in the sixth version. 
When we train ScaleDet on these four datasets, there are 2249 class labels in the unified label space (Eq.~\eqref{eq:concat}) and 3.7M training images. 

\myparagraph{Downsteam datasets.} To evaluate the transferability on diverse unseen datasets, we adopt the recent challenging ``Object Detection in the Wild'' (ODinW) benchmark \cite{li2022grounded}. It contains 13 public object detection datasets with very different application domains, to mimic diverse and challenging real-world scenarios. 
Some of these datasets capture seen classes in unseen domains, like people in thermal images, while others capture unseen classes, like ``jellyfish'' and ``stingray'' in aquariums (Figure \ref{fig:teaser}, bottom right). 
To evaluate on ODinW, the models can be evaluated directly (direct transfer) or evaluated after fine-tuning (fine-tune transfer). 

\myparagraph{Evaluation metrics.} 
For evaluation on {\it upstream} datasets, we use the standard mAP metric (i.e., mAP at IoU thresholds 0.5 to 0.95) on COCO, LVIS and Objects365. On OpenImages, we follow the official evaluation protocol that uses mAP@0.5 and an expanded semantic class hierarchy \cite{kuznetsova2020open}.
For evaluation on {\it downstream} datasets, we follow \cite{li2022grounded} and report the average mAP over 13 datasets in this section. We provide the detailed qualitative and quantitative results on individual datasets in {\em Supplementary}. 

\myparagraph{Implementation details.} 
In our experiments, unless explicitly stated, we use CenterNet2 \cite{zhou2021probabilistic} with a backbone pre-trained on ImageNet21k \cite{ridnik2021imagenet}. We use prompt text embeddings from CLIP \cite{radford2021learning} or OpenCLIP \cite{ilharco_gabriel_2021_5143773} to encode class labels. For augmentation, we use large scale jittering \cite{ghiasi2021simple} and efficient resize crop \cite{zhou2022detecting} with an input size of $640{\times}640$, $896{\times}896$ when using ResNet50 \cite{he2016deep}, Swin Transformer \cite{liu2021swin} as backbone. We use an input size of $800{\times}1333$ for testing. We use Adam optimizer \cite{kingma2014adam} and train on 8 V100 GPUs. For multi-dataset training, we directly combine all datasets and use repeat factor sampling \cite{gupta2019lvis}, without using any   multi-dataset sampling strategy. More details about learning schedules of different tables are in {\em Supplementary}. 

\subsection{Training with a growing number of datasets}
\label{sec:analysis}

\myparagraph{Evaluation setup.} 
We setup the single-dataset {\it baseline} by training on individual datasets. We use text embeddings to represent the class labels for each dataset, and train the detector on each dataset by hard label assignment (Eq.~\eqref{eq:hard}). 
To evaluate how ScaleDet scales as we increase the number of training datasets, we train it using the following incremental combinations: (1) LVIS+COCO, (2) LVIS+COCO+O365, (3) LVIS+COCO+O365+OID. These increase the dataset size from $218k \rightarrow 1.96M \rightarrow 3.7M$ and the number of class labels from $1283 \rightarrow 1648 \rightarrow 2249$. 
After training the baseline and ScaleDet, we replace the text embeddings in Eq.~\eqref{eq:cos} to represent the labels of each dataset for evaluation. For example, to evaluate on COCO we set the text embeddings to represent the 80 classes of COCO and derive class predictions based on visual-textual similarities.

\begin{figure}[!t]
  \centering 
   \includegraphics[width=0.82\linewidth]{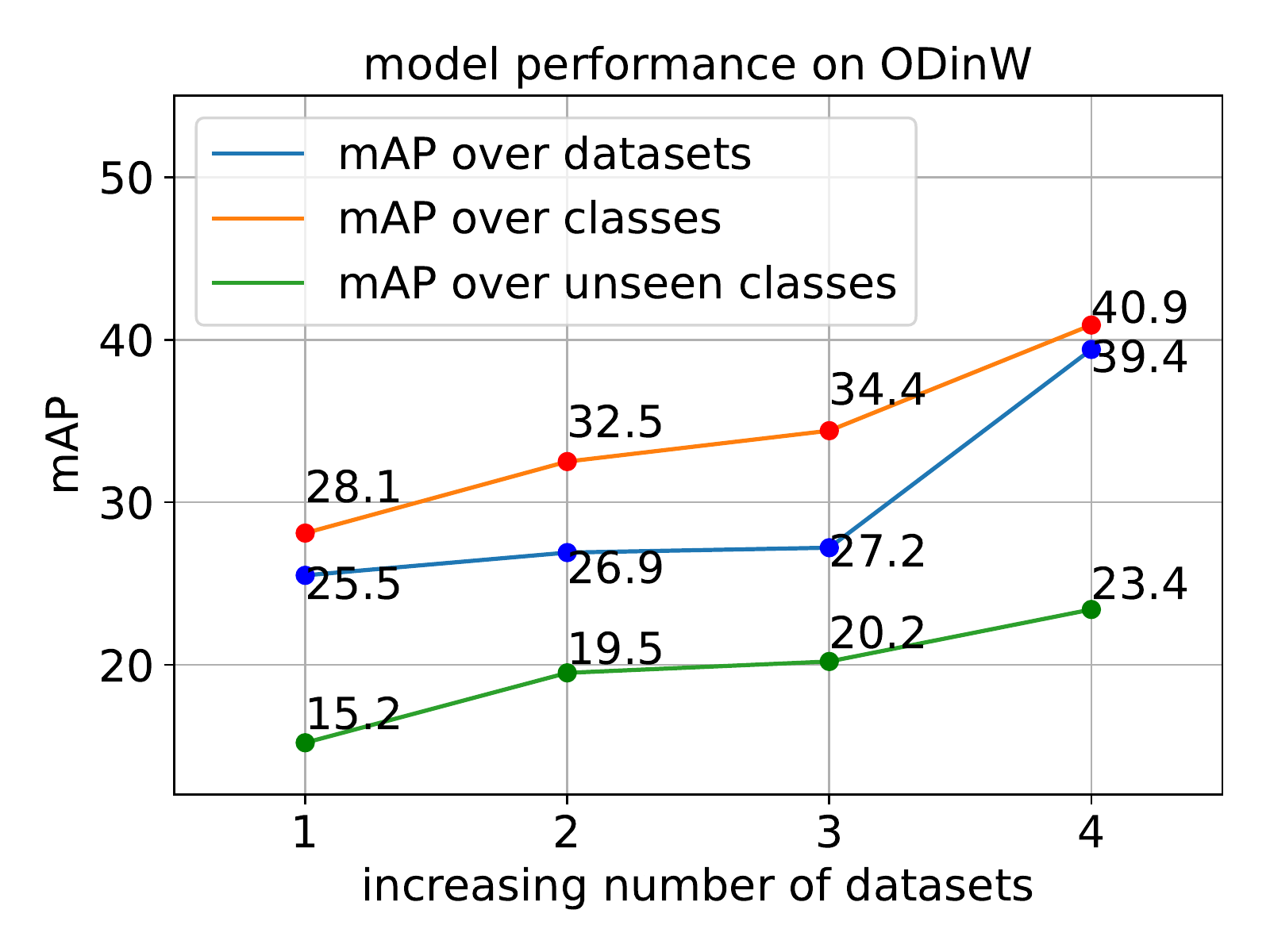}
   \vskip -1em
   \caption{{\bf Direct transfer results on ODinW benchmark} when scaling up the number of training datasets. 
   Number 1-4 in $x$-axis means using L, L+C, L+C+O365, L+C+O365+OID for training. 
   Details of unseen classes, qualitative results are in {\em Supplementary}. 
   }
   \label{fig:increasing1}
   \vskip -1.5em
\end{figure}

\begin{figure}[!t]
  \centering
   \includegraphics[width=0.9\linewidth]{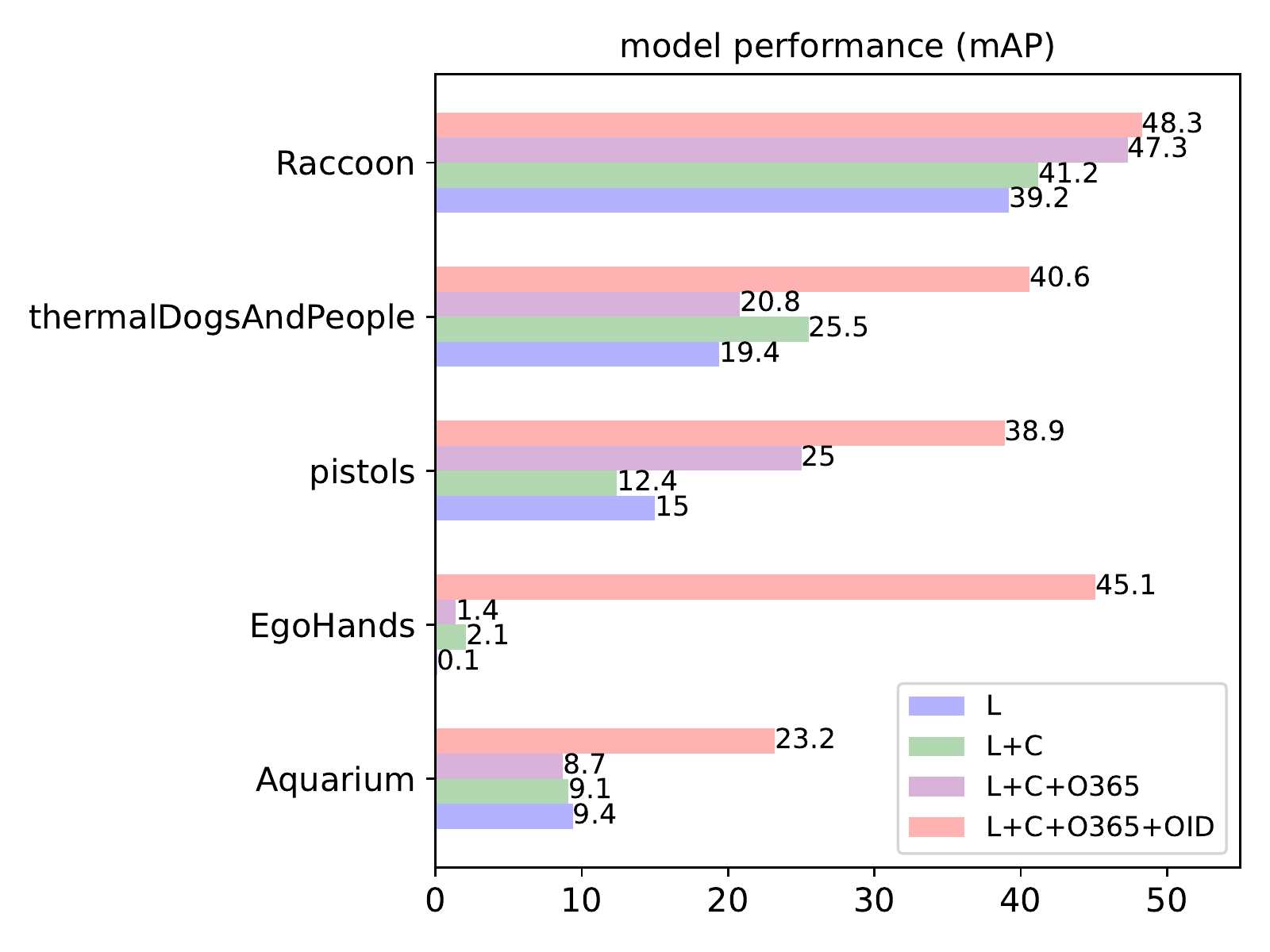}
   \vskip -1em
   \caption{{\bf Direct transfer to different datasets on ODinW benchmark} when scaling up the number of training datasets. 
   }
   \label{fig:increasing2}
   \vskip -1.5em
\end{figure}

To make the evaluation efficient, for the results in Table~\ref{tab:increasing}, Figure~\ref{fig:increasing1} and Figure~\ref{fig:increasing2} we use a fixed input size of 640$\times$640, a ResNet50 \cite{he2016deep} backbone and text embeddings from CLIP \cite{radford2021learning}. 
We then explore using larger backbones (Swin Transformers \cite{liu2021swin}) and text embeddings from OpenCLIP \cite{ilharco_gabriel_2021_5143773} for the analysis of backbones in Table~\ref{tab:backbone}. 

\begin{table}[!t]
  \centering
  \resizebox{0.94\linewidth}{!}{%
  \begin{tabular}{l @{\hskip 1.5mm} l @{\hskip 1.5mm} | @{\hskip 1.5mm} l @{\hskip 1.5mm} l @{\hskip 1.5mm} l @{\hskip 1.5mm} l @{\hskip 1.5mm} | @{\hskip 1.5mm} l}
    \toprule
    Model & Dataset(s) & L & C & O365 & OID & mAP \\
    \midrule
    \multirow{4}{*}{baseline} 
    & L & \color{green}33.1 & 37.0 & 15.2 & 41.5 & 31.7 \\
    & C & 11.0 & \color{green}46.8 & 7.9 & 33.1 & 24.7 \\ 
    & O365 & 19.2 & 39.8 & \color{green}28.8 & 47.6 & 33.9 \\
    & OID & 15.7 & 31.3 & 14.1 & \color{green}69.3 & 32.6 \\
    \midrule 
    \multirow{3}{*}{\bf ScaleDet} 
    & L,C & 33.3 & 44.9 & 15.9 & 43.7 & 34.5 \\
    & L,C,O365 & 36.5 & 47.0 & \bf 31.2 & 44.9 & 39.9 \\
    & L,C,O365,OID & \bf 36.8 & \bf 47.1 & 30.6 & \bf 69.4 & \bf 46.0 \\
    \bottomrule
  \end{tabular}
  }
  \vskip -0.5em
  \caption{{\bf Training with an increasing number of datasets.} The best results on single-/multi-dataset training are in {\color{green}green}/{\bf bold}. Backbone: ResNet50. 
  mAP is the mean AP across datasets. 
  }
\label{tab:increasing}
\vskip -1em
\end{table} 
\myparagraph{Effect on upstream datasets.} 
Table~\ref{tab:increasing} shows the evaluation on upstream datasets when increasing the number of datasets, from which we can observe the followings. {\em First}, scaling up the number of training datasets consistently leads to better model performance, e.g., the mAP on LVIS is improved from 33.1, 33.3, 36.5, to 36.8 when increasing the number of training sets from 1 to 4. 
{\em Second}, multi-dataset training with ScaleDet is generally better than single-dataset training (baseline), e.g., the mAP on LVIS, COCO, O365, OID are {greatly} improved from 33.1, 46.8, 28.8, 69.3 (baseline) to 36.8, 47.1, 30.6, 69.4 when training on all datasets. 
This suggests that ScaleDet learns well across heterogeneous label spaces, diverse domains of different datasets, and does not overfit to any specific datasets.

\myparagraph{Effect on downstream datasets.}  
Figure~\ref{fig:increasing1} shows the results of direct transfer on the ODinW benchmark. 
Notably, scaling up the number of training datasets for ScaleDet significantly improves its accuracy (mAP) on downstream datasets, for all settings (i.e., over datasets, over classes and over unseen classes).
In Figure~\ref{fig:increasing2} we further visualize the performance of ScaleDet on some of the downstream datasets in ODinW. These datasets either contain unseen classes or come from the visual domains that are very different from those used for training. Importantly, ScaleDet performs well on both scenarios. For example, {\it thermalDogsAndPeople} contains novel thermal images, yet multi-dataset training with ScaleDet is capable of boosting its performance considerably from 19.4 mAP (single-dataset) to 40.6 (all 4). Furthermore, {\it Raccoon} requires the localization of an unseen class, yet again ScaleDet improves zero-shot detection from 39.2 mAP (single-dataset) to 48.3 (all 4). 
Overall, Figure~\ref{fig:increasing1} and Figure~\ref{fig:increasing2} show how ScaleDet can utilize the synergistic benefits of different upstream datasets to boost its generalization on downstream datasets.

\begin{table}[!t]
  \centering
  \resizebox{0.94\linewidth}{!}{%
  \begin{tabular}{l @{\hskip 1mm}|@{\hskip 1mm} l @{\hskip 1mm} l @{\hskip 1mm} | @{\hskip 1mm} l @{\hskip 1mm} l @{\hskip 1mm} l @{\hskip 1mm} l | @{\hskip 1mm} l}
    \toprule
    &Model & Text & L & C & O365 & OID & mAP \\
    \midrule
   1&ScaleDet-R & CLIP & 36.8 & 47.1 & 30.6 & 69.4 & 46.0 \\
   \midrule
   2&ScaleDet-T& CLIP & 42.7 & 51.2 & 35.9 & 73.6 & 50.9 \\
   3&ScaleDet-T& OpenCLIP & 43.9 & 51.4 & 36.3 & 74.1 & 51.4 \\
   \midrule
   4&ScaleDet-B & CLIP & 50.4 & 55.6 & 43.7 & 75.7 & 56.3 \\
   5&ScaleDet-B & OpenCLIP & \bf 50.7 & 55.4 & 43.8 & \bf 76.2 & 56.5 \\
   6&ScaleDet-B$^ *$ & OpenCLIP & 50.6 & \bf 58.8 & \bf 46.8 & 75.9 & \bf 58.0 \\
    \bottomrule
  \end{tabular}
  }
  \vskip -0.5em
  \caption{{\bf Different backbones for multi-dataset training.}
  R: ResNet50.
  T, B: Swin-Tiny, Swin-Base transformer. 
  $^ *$: further fine-tune the model on each dataset after multi-dataset training. 
  }
\label{tab:backbone}
\vskip -1em
\end{table}

\myparagraph{Effect of different backbones.} 
We now test ScaleDet with different backbones and text embeddings (Table~\ref{tab:backbone}). Enhancing the discriminative power of the image backbone from ResNet50 \cite{he2016deep} to Swin-Base Transformer \cite{liu2021swin} significantly boosts the mAP from 46.0 to 56.3 (row 1 vs 4). Changing the text embeddings from CLIP to OpenCLIP instead helps mostly when using weaker backbones (e.g., +0.5 mAP with ScaleDet-T, row 2 vs 3), as opposed to stronger ones (e.g., +0.2 mAP with ScaleDet-B, row 4 vs 5). Finally, with fine-tuning ScaleDet can be further improved from 56.5 to 58.0 in mAP (row 5 vs 6).
Overall, all these results suggest that ScaleDet can leverage the complementary power of scalable training sets and strong backbones to achieve competitive model performance. 

\subsection{Comparison to SOTA multi-dataset detectors}
\label{sec:multisota}

In this section, we provide an apple-to-apple comparison of ScaleDet against the two latest state-of-the-art (SOTA) multi-dataset detectors: UniDet \cite{zhou2022simple} and Detic \cite{zhou2022detecting}. 

\myparagraph{Comparison to UniDet.} 
In Table~\ref{tab:unidet}, we follow the settings of UniDet \cite{zhou2022simple} and train ScaleDet on the same datasets (COCO, O365, OID), using the same ResNet50 backbone, SGD optimizer and standard data augmentation in the Detectron2 codebase \cite{wu2019detectron2}. In UniDet, multiple dataset-specific classifiers are trained, while ScaleDet is trained with semantic labels by one classifier. Table~\ref{tab:unidet} shows the comparison on multi-dataset training. ScaleDet outperforms UniDet, yielding an mAP of 47.7 vs 45.4. Moreover, ScaleDet has better improved margins for multi-dataset training over single-dataset training (row 4,3 vs 2,1). The improved margin in mAP is 2.6 points (47.7 vs 45.1) by ScaleDet vs 1.1 points (45.4 vs 44.3) by UniDet. These results show the benefit of learning in a unified semantic label space, as opposed to training multiple dataset-specific classifiers (UniDet). 

\begin{table}[!t]
  \centering
\resizebox{0.95\linewidth}{!}{%
\begin{tabular}{l @{\hskip 1mm}|@{\hskip 1mm} l@{\hskip 1mm}l @{\hskip 1mm}| @{\hskip 1.5mm} ccc @{\hskip 2mm}| @{\hskip 2mm}l}
    \toprule
    & Model & Dataset(s) & COCO & O365 & OID & mAP \\ \midrule
    1&\multirow{2}{*}{UniDet} & single & 42.5 & 24.9 & 65.7 & 44.3 \\
    2&& multiple & \bf 45.5 & 24.6 & 66.0 & 45.4 \\
    \midrule
    3&\multirow{2}{*}{\bf ScaleDet} & single & 42.1 & 26.5 & 66.6 & 45.1 \\
    4&& multiple & \bf 45.5 & \bf 27.9 & \bf 69.6 & \bf 47.7 \\
    \bottomrule
  \end{tabular}
  }
  \vskip -0.5em
  \caption{{\bf Comparison to UniDet \cite{zhou2022simple} on multi-dataset training} with COCO, O365 and OID. Backbone: ResNet50 pre-trained on ImageNet1k.  
  ``single'': training with one single training set same as the test set. ``multiple'': multi-dataset training with COCO,O365,OID. mAP is the mean AP across datasets.
  }
  \label{tab:unidet}
  \vskip -1em
\end{table}

\begin{table}[!t]
  \centering
  \resizebox{0.95\linewidth}{!}{%
  \begin{tabular}{l @{\hskip 1.5mm}|@{\hskip 1.5mm} l @{\hskip 2mm} l @{\hskip 1.5mm} | @{\hskip 1.5mm} cc | l}
    \toprule
    & Model & Datasets & LVIS & COCO & mAP \\ 
    \midrule
    1& Detic \cite{zhou2022detecting} & L,C & 33.0 & 43.9 & 38.4 \\
    2& \bf ScaleDet & L,C & \bf 33.3 & \bf 44.9 & \bf 39.1 \\
    \midrule
    3& Detic \cite{zhou2022detecting} & L,C,IN21k & 35.4 & 42.4 & 38.9 \\
    4& \bf ScaleDet & L,C,O365 & 36.5 & 47.0 & 41.7  \\
    5& \bf ScaleDet & L,C,O365,OID & \bf 36.8 & \bf 47.1 & \bf 41.9 \\
    \bottomrule
  \end{tabular}
  }
  \vskip -0.5em
  \caption{{\bf Comparison to Detic \cite{zhou2022detecting} on multi-dataset training} with LVIS (L) and COCO (C). Backbone: ResNet50. IN21k: ImageNet21k \cite{deng2009imagenet}. In row 1-2, both Detic and ScaleDet are trained with only LVIS and COCO, thus providing a like-to-like comparison. In row 3, 4, 5, the training data size is 12.6M, 1.96M, 3.7M respectively. mAP is the mean AP across datasets. 
  }
  \label{tab:detic}
  \vskip -1em
\end{table}
\myparagraph{Comparison to Detic.} In Table~\ref{tab:detic}, we follow the settings of Detic \cite{zhou2022detecting} and perform multi-dataset training on LVIS and COCO, using a ResNet50 backbone. In Detic, the unified label space for LVIS and COCO contains 1203 class labels, obtained by merging two label sets with {\em wordnet synsets}, while ScaleDet `flattens' their labels (1203+80) to 1283. 
Table \ref{tab:detic} (row 1 and 2) presents the comparison. ScaleDet outperforms Detic by 1 point on COCO (44.9 \vs 43.9) and 0.7 mAP on average.
Next, we compare ScaleDet and Detic with more training sets (Table \ref{tab:detic}, row 3-5). ScaleDet is trained using more detection datasets (O365,OID), while Detic uses 14M additional classification images (ImageNet21) as a large-scale weakly annotated dataset for detection. 
Our results show that, although Detic is trained with far more data and class labels than ScaleDet (12.6M images and 22k classes after data cleaning vs 3.7M images and 2k classes), ScaleDet still surpasses Detic's performance by a large margin of 3.0 mAP points (41.9 \vs 38.9). 
{These results show the importance of learning from multiple detection datasets and the effectiveness of ScaleDet in doing so.} 

\subsection{Comparison to SOTA detectors on COCO}
\label{sec:sota}

\begin{table}[!t]
  \centering
  \resizebox{0.82\linewidth}{!}{%
  \begin{tabular}{l | l | l | l}
    \toprule
     & Model & Model Type & mAP \\
    \midrule
     1  & Faster RCNN \cite{ren2015faster} & \multirow{6}{*}{\makecell[l]{single-dataset \\ detection}} & 37.9 \\
     2  & Mask RCNN \cite{he2017mask} & & 39.8 \\
     3  & CenterNet \cite{zhou2019objects} & & 40.2 \\ 
     4  & CascadeRCNN \cite{cai2018cascade} & & 41.6 \\
     5  & DETR \cite{carion2020end} & & 42.0 \\
     6  & CenterNet2 \cite{zhou2021probabilistic} & & 42.9 \\
     \midrule
     7 & UniT \cite{hu2021unit} & \multirow{2}{*}{\makecell[l]{detection +  \\ understanding}} & 42.3 \\
     8 & RegionCLIP \cite{zhong2022regionclip} & & 42.7 \\
     \midrule
     9 & Detic \cite{zhou2022detecting} & \makecell[l]{detection +  \\ classification} & 42.4 \\
     \midrule
     10 & UniDet \cite{zhou2022simple} & \multirow{2}{*}{\makecell[l]{multi-dataset \\ detection}} & 45.5 \\
     11 & \bf ScaleDet & & \bf 47.1 \\
    \bottomrule
  \end{tabular}
  }
  \vskip -0.5em
  \caption{{\bf Comparison on COCO using ResNet50 as backbone.} Note: ``understanding'', ``classification'' mean training with vision-and-language understanding, and classification datasets respectively. 
  All methods include COCO dataset for training.
  }
  \label{tab:coco1}
  \vskip -1em
\end{table}

\begin{table}[!t]
  \centering
  \resizebox{0.82\linewidth}{!}{%
  \begin{tabular}{l | l | l | l}
    \toprule
     & Model & Model Type & mAP \\
    \midrule
     1  & Faster RCNN-T \cite{ren2015faster} & \multirow{3}{*}{\makecell[l]{single-dataset \\ detection}} & 46.0 \\
     2  & DyHead-T \cite{dai2021dynamic} & & 49.7 \\
     3  & CascadeRCNN-T \cite{cai2018cascade} & & 50.4 \\ 
     \midrule
     4 & GLIP-T \cite{li2022grounded} & \multirow{3}{*}{\makecell[l]{detection + \\ understanding}} & 55.2 \\
     5 & GLIPv2-T \cite{zhang2022glipv2} & & 55.5 \\  
     6 & GLIPv2-B \cite{zhang2022glipv2} & & \bf 58.8 \\
     \midrule
     7 & Detic-B \cite{zhou2022detecting} & \makecell[l]{detection +  \\ classification} & 54.9 \\
     \midrule
     8 & \bf ScaleDet-B & {\makecell[l]{multi-dataset \\ detection}} & \bf 58.8 \\
    \bottomrule
  \end{tabular}
  }
  \vskip -0.5em
  \caption{{\bf Comparison on COCO using Swin Transformer as backbone.} 
  T, B: Swin-Tiny, Swin-Base Transformer. 
  Note: GLIPv2-B uses 20.5M data: 3.7M detection data (COCO, O365, OID, Visual genome \cite{krishna2017visual}, ImageNetBoxes \cite{deng2009imagenet}), 16.8M grounding and caption data (GoldG \cite{kamath2021mdetr}, CC15M \cite{sharma2018conceptual,changpinyo2021conceptual}, SBU \cite{ordonez2011im2text}); while
  ScaleDet-B uses 3.5M data (LVIS,COCO,O365,OID).
  }
  \label{tab:coco2}
  \vskip -1em
\end{table}

In this section, we compare ScaleDet to different types of SOTA detectors on standard COCO benchmark.
The different models cover four major types: (1) single-dataset detectors; (2) detectors trained with vision-and-language understanding datasets, e.g., UniT \cite{hu2021unit} is trained on 7 tasks over 8 datasets and RegionCLIP \cite{zhong2022regionclip} is trained with 3M image-text pairs from Conceptual Caption \cite{sharma2018conceptual}; (3) detector trained with image classification dataset, i.e., Detic \cite{zhou2022detecting} is trained with ImageNet21k \cite{deng2009imagenet}; (4) multi-dataset detector trained with only detection datasets, e.g., UniDet \cite{zhou2022simple}. 

We train our ScaleDet using LVIS, COCO, O365, OID and present the results in Table~\ref{tab:coco1}, where all models are trained using a ResNet50 backbone. ScaleDet obtains the best performance among 11 models. 
Moreover, compared to detectors trained with large vision-and-language or classification datasets (row 7-9), ScaleDet gives much better performance, even though it learns a much smaller number of concepts -- ScaleDet achieves 47.1 by "only" learning 2249 labels, while RegionCLIP and Detic achieve 42.7 and 42.4 by learning 6790 and 22047 concepts, respectively. 

In Table~\ref{tab:coco2}, we compare ScaleDet to competitors using Swin Transformers as backbone. 
With this strong backbone, ScaleDet demonstrates SOTA performance with high data-efficiency among all models. Among all competitors, the recent GLIPv2-B \cite{zhang2022glipv2} is the only model performing on-par with ScaleDet (58.8), but using almost an order of magnitude more training data than ScaleDet (20.5M \vs 3.5M). 

Overall, our results in Table~\ref{tab:coco1} and Table \ref{tab:coco2} show that ScaleDet offers a data-efficient training recipe that can learn from {fewer visual concepts and less training data, but still achieves} state-of-the-art performance on COCO.

\subsection{Comparison of SOTA on ODinW}
\label{sec:odinw}

\begin{table}[!t]
  \centering
  \resizebox{0.94\linewidth}{!}{%
  \begin{tabular}{l@{\hskip 1mm}|@{\hskip 1mm}l@{\hskip 1mm}|@{\hskip 1mm}l@{\hskip 1mm}|@{\hskip 1mm}x{1.3cm}@{\hskip 1mm}x{1.3cm}}
    \toprule
    \multirow{2}{*}{Model} & \multirow{2}{*}{Model Type} & \multirow{2}{*}{\#Data} & \multicolumn{2}{c}{ODinW} \\
    & & & {direct} & {fine-tune} \\
    \midrule
    GLIP-T\cite{li2022grounded} & \multirow{3}{*}{\makecell[l]{detection +\\  understanding}} & 5.5M & 46.5 & 64.9 \\
    GLIPv2-T\cite{zhang2022glipv2} & & 5.5M & 48.5 & 66.5 \\ 
    GLIPv2-B\cite{zhang2022glipv2} & & 20.5M & \bf 54.2 & 69.4 \\
    \midrule
    Detic-R\cite{zhou2022detecting} & \multirow{2}{*}{\makecell[l]{detection +\\  classification}} & 12.6M & 29.4 & 64.4 \\
    Detic-B\cite{zhou2022detecting} & & 12.6M & 38.7 & 70.1 \\
    \midrule
    ScaleDet-R & \multirow{3}{*}{\makecell[l]{detection}} & 3.6M & 39.4 & 68.5 \\
    ScaleDet-T & & 3.6M & 44.3 &  70.4 \\ 
    ScaleDet-B & & 3.6M & 47.3 & \bf 71.8 \\
    \bottomrule
  \end{tabular}
  }
  \vskip -0.5em
  \caption{{\bf Results of direct and fine-tune transfer on ODinW.} 
  R, T, B: ResNet50, Swin-Tiny, Swin-Base Transformer as backbone. 
  Metric: mAP over 13 datasets. Note: ``understanding'', ``classification'' mean training with vision-and-language understanding, and classification datasets besides detection data.
  }
  \label{tab:odinw}
\vskip -1em
\end{table}

We evaluate the transferablity on the ``Object Detection in the Wild'' (ODinW) benchmark and compare ScaleDet with SOTA pre-trained detectors that are capable of both direct and fine-tune transfer given any downstream detection datasets.  
Table~\ref{tab:odinw} shows the comparison on ODinW among 3 types of detectors: (1) GLIP \cite{li2022grounded}, GLIPv2 \cite{zhang2022glipv2}, (2) Detic \cite{zhou2022detecting}, and (3) ScaleDet. 
ScaleDet is trained with the smallest training data, yet it achieves the best results in fine-tune transfer, surpassing even GLIPv2-B by 2.4 points (71.8 \vs 69.4). On direct transfer, GLIPv2-B is however stronger; we conjecture that this is thanks to its massive visual-language training data (20.5M), which likely covers the unseen concepts in the downstream datasets.
Furthermore, when comparing ScaleDet-R, ScaleDet-B to Detic-R, Detic-B, we find the results of ScaleDet are significantly better, e.g., ScaleDet-R outperforms Detic-R by 10.0 points (39.4 \vs 29.4) on direct transfer and by 4.1 (68.5 \vs 64.4) on fine-tune transfer. 
These observations show the efficacy of ScaleDet in transferring to downstream datasets.

\subsection{Ablation study}
\label{sec:ablation}

\begin{table}
\begin{subtable}[t]{0.48\textwidth}
\centering
\resizebox{0.85\linewidth}{!}{
\begin{tabular}{l@{\hskip 1mm}l @{\hskip 1mm}|@{\hskip 1.5mm} l@{\hskip 1mm}l@{\hskip 1mm}l@{\hskip 1mm}l @{\hskip 1.5mm}| @{\hskip 1.5mm}l}
    \toprule
    Model & Datasets & L & C & O365 & OID & mAP \\
    \midrule
    $\mathcal{L}_{hl}$ & \multirow{2}{*}{L,C} & 
    33.1 & 44.6 & 15.7 & 43.2 & 34.2
    \\
    $\mathcal{L}_{hl}{+}\mathcal{L}_{sl}$  & &
    33.3 & 44.9 & 15.9 & 43.7 & \bf 34.5
    \\
    \midrule 
    $\mathcal{L}_{hl}$ & \multirow{2}{*}{L,C,O365,OID} & 
   36.7 & 46.7 & 30.8 & 69.1 & 45.8
    \\
    $\mathcal{L}_{hl}{+}\mathcal{L}_{sl}$ & &
   36.8 & 47.1 & 30.6 & 69.4 & \bf 46.0
    \\
    \bottomrule
\end{tabular}
}
\caption{Ablation on upstream datasets.}
\label{tab:ablation1}
\end{subtable}

\hspace{\fill}

\begin{subtable}[t]{0.48\textwidth}
  \centering
  \resizebox{0.72\linewidth}{!}{
  \begin{tabular}{l l | l l}
    \toprule
    \multirow{2}{*}{Model} & Datasets &  \multicolumn{2}{c}{ODinW} \\  \cmidrule{2-4}
    & Metric & mAP & mAP${_C}$ \\
    \midrule
    $\mathcal{L}_{hl}$ & \multirow{2}{*}{L,C} & 
    26.9 & 32.1 \\
    $\mathcal{L}_{hl}{+}\mathcal{L}_{sl}$  & &
    \bf 26.9 & \bf 32.5 \\
    \midrule 
    $\mathcal{L}_{hl}$ & \multirow{2}{*}{L,C,O365,OID} & 
   37.9 & 39.8  \\
    $\mathcal{L}_{hl}{+}\mathcal{L}_{sl}$ & &
    \bf 39.4 & \bf 40.9 \\
    \bottomrule
  \end{tabular}
  }
  \caption{Ablation on downstream benchmark ODinW by direct transfer.}
  \label{tab:ablation2}
\end{subtable}

\vskip -0.7em
\caption{Ablation study on upstream datasets \subref{tab:ablation1} and downstream datasets \subref{tab:ablation2}. mAP is the mean AP over datasets; mAP${_C}$ is the mean AP over all classes; better mAP of each setup is in {\bf bold}.}
\vskip -1em
\end{table} 

Finally, we ablate the components of our ScaleDet. As described in Sec.~\ref{sec:method}, ScaleDet is trained with two loss terms: the hard label assignment $\mathcal{L}_{hl}$ (Eq.~\eqref{eq:hard}), and the soft label assignment $\mathcal{L}_{sl}$ (Eq.~\eqref{eq:soft}). $\mathcal{L}_{hl}$ aims to assign each visual feature to one corresponding label, while $\mathcal{L}_{sl}$ works as a regularizer to relate labels across datasets. 
Table~\ref{tab:ablation1} shows our ablation study on 4 upstream datasets and Table~\ref{tab:ablation2} shows our ablation of direct transfer on downstream benchmark ODinW. 
We find that training with two loss terms ($\mathcal{L}_{hl}{+}\mathcal{L}_{sl}$) leads to better overall results on both upstream and downstream datasets. For instance, in Table \ref{tab:ablation1}, when training with two datasets (L+C), $\mathcal{L}_{hl}{+}\mathcal{L}_{sl}$ 
{gives consistent better results on 4 upstream datasets compared to $\mathcal{L}_{hl}$.} 
In Table \ref{tab:ablation2}, when using $\mathcal{L}_{hl}{+}\mathcal{L}_{sl}$ to train on four training sets (L+C+O365+OID), the mAP is greatly improved by 1.5 points (39.4-37.9) and mAP${_C}$ is improved by 1.1 points (40.9-39.8), as compared to using only $\mathcal{L}_{sl}$.
This suggests that $\mathcal{L}_{sl}$ is valuable to improve the model's generalizability.
{Overall, these results indicate that $\mathcal{L}_{hl}$ works effectively for label assignment within the upstream dataset, while $\mathcal{L}_{sl}$ takes a complementary role that is more evident when transferring to downstream datasets.}

\section{Conclusion and Future Work}
\label{sec:conclusion}

We present ScaleDet, a simple yet scalable and effective multi-dataset training recipe for detection. ScaleDet learns across multiple datasets in a unified semantic label space, optimized by hard label and soft label assignments to align visual and text embeddings. ScaleDet achieves the new state-of-the-art performance on multiple upstream datasets (LVIS, COCO, Objects365, OpenImages) and downstream datasets (ODinW). As the key success of ScaleDet lies in learning in a unified semantic label space, our formulation can also generalize to other vision tasks such as image classification and semantic segmentation. We leave the unified design of a generic multi-dataset multi-task foundation model as promising and exciting future endeavors.

\section*{\Large Supplementary Material}
\setcounter{figure}{0}
\setcounter{section}{0}
\setcounter{table}{0}
\setcounter{algorithm}{0}
\renewcommand\thesection{\Alph{section}}
\renewcommand\thefigure{\Alph{figure}}
\renewcommand\thetable{\Alph{table}}
\renewcommand\thealgorithm{\Alph{algorithm}}

\section{Additional results} 
\label{sec:results}

\myparagraph{Additional results on O365 and OID.}
In Sec 4.3 Table 4, we follow the evaluation of Detic \cite{zhou2022detecting} and provide results on two training datasets: LVIS and COCO. As Detic is trained with text embeddings, it can also be tested on other datasets. In Table \ref{tab:detic}, we compare ScaleDet and Detic on two additional datasets: Objects365 (O365), OpenImages (OID), along with LVIS and COCO, where both models use Swin-Base Transformer as backbone. For Detic, we use the best model ``Detic\_C2\_SwinB\_896\_4x\_IN-21K+COCO'' from the Detic model zoo\footnote{\url{https://github.com/facebookresearch/Detic/blob/main/docs/MODEL_ZOO.md}}. For ScaleDet, we use our best model in Table 2 in Sec 4.2. In Table \ref{tab:detic}, it can be seen that our ScaleDet surpasses Detic with substantial margins, improving the mAP over all datasets from 42.7 to 56.6. 

\begin{table}[!h]
  \centering
  \resizebox{0.92\linewidth}{!}{%
  \begin{tabular}{llllll}
    \toprule
    Model & LVIS & COCO & OID & O365 & mAP \\ 
    \midrule
    Detic-B \cite{zhou2022detecting} &  44.3 & 50.1 & 21.6 & 54.6 & 42.7 \\
   \bf ScaleDet-B & \bf 50.7 & 55.4 & 43.8 & \bf 76.2 & 56.5 \\
   \bf ScaleDet-B$^ *$ & 50.6 & \bf 58.8 & \bf 46.8 & 75.9 & \bf 58.0 \\
    \bottomrule
  \end{tabular}
  }
  \vskip -1em
  \caption{{\bf Comparison to Detic \cite{zhou2022detecting} on multi-dataset training} using the best models from both papers. B: using Swin-Base Transformer as backbone. 
  $^ *$: model is further fine-tuned on each dataset after multi-dataset training.
  mAP is the mean AP over all datasets. 
  }
  \label{tab:detic}
  \vskip -1em
\end{table}

\myparagraph{Results of individual datasets on ODinW.} In Table \ref{tab:odinw13}, 
we provide the results on ``Object Detection in the Wild'' benchmark, which reports the mAP on individual datasets: AerialDrone, Aquarium, Rabbits, EgoHands, Mushrooms, Packages, PascalVOC, pistols, pothole, Raccoon, Shellfish, Thermal, Vehicles, as well as the mAP over 13 datasets. 

\begin{figure*}[!ht]
  \centering
   \includegraphics[width=0.96\linewidth]{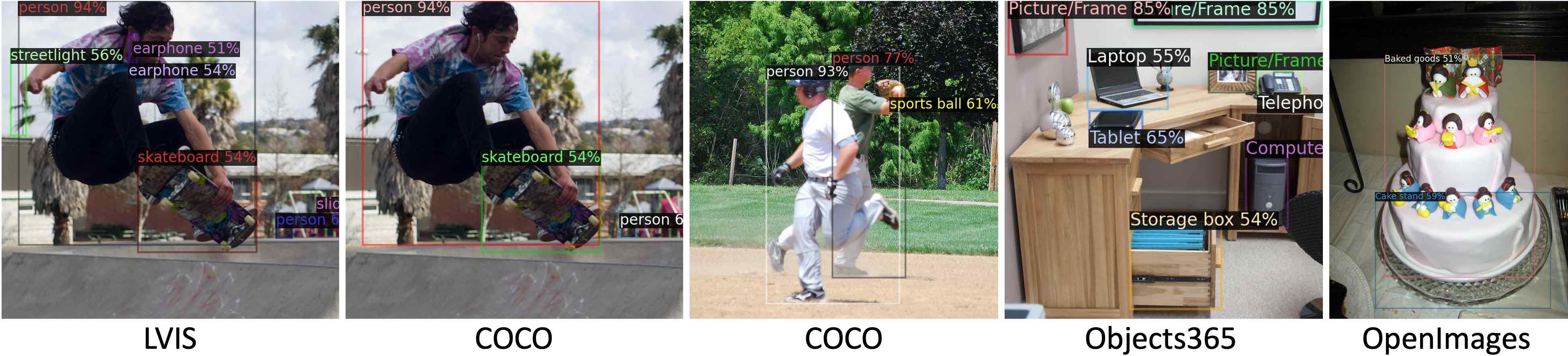}
   \vskip -0.7em
   \caption{
  {\bf  Qualitative results testing example images on upstream datasets}: LVIS, COCO, Objects365 and OpenImages. It can be seen that on the same image (in column 1 and 2), the model can detect different classes. In different columns, the model is evaluated by setting the classifier to recognize a certain set of class labels from one dataset at test time.
   }
   \label{fig:upstream}
\end{figure*}

\begin{figure*}[!ht]
  \centering
   \includegraphics[width=0.96\linewidth]{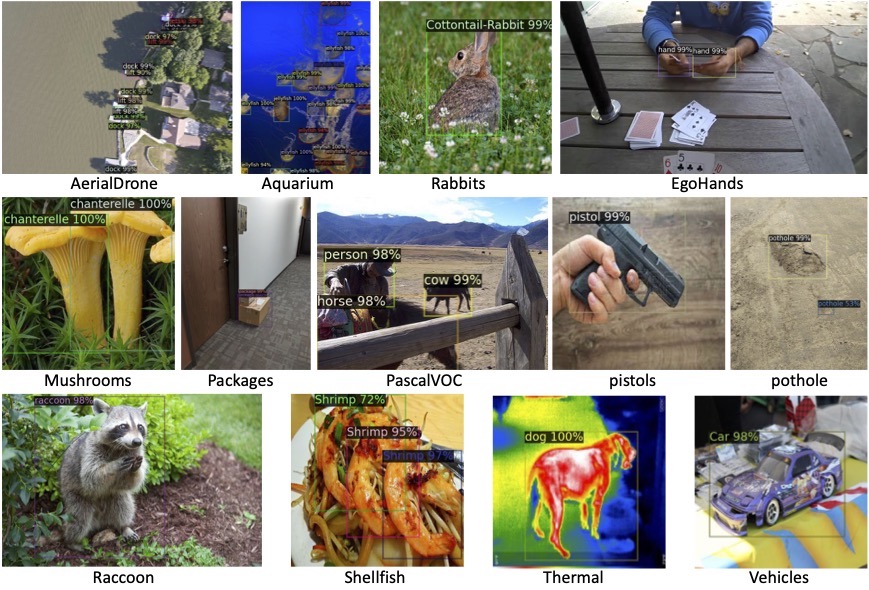}
   \vskip -0.7em
   \caption{
  {\bf  Qualitative results testing example images on downstream datasets}: AerialDrone, Aquarium, Rabbits, EgoHands, Mushrooms, Packages, PascalVOC, pistols, pothole, Raccoon, Shellfish, Thermal, Vehicles.
   }
   \label{fig:downstream}
\end{figure*}

\myparagraph{Qualitative results.}
Due to space limit in our manuscript, we provide additional qualitative results on upstream and downstream datasets in Figure \ref{fig:upstream} and Figure \ref{fig:downstream} respectively.

\section{Implementation details} 
\label{sec:implement}

 \begin{table}[!ht]
   \centering
   \resizebox{\linewidth}{!}{%
   \begin{tabular}{ll}
     \toprule
     Dataset & Website \\
     \midrule
     LVIS & \url{https://www.lvisdataset.org} \\
     COCO & \url{http://cocodataset.org} \\ 
     Objects365 & \url{https://www.objects365.org} \\
     OpenImages & \url{https://storage.googleapis.com/openimages/web/index.html} \\
     \bottomrule
   \end{tabular}
   }
   \vskip -1em
   \caption{Dataset details on 4 upstream datasets. All datasets are under the ``CC BY 4.0'' license.}
 \label{tab:upstream}
  \vskip -1em
 \end{table}

\myparagraph{Dataset details.} 
We provide the licenses and websites of upstream and downstream datasets in Table \ref{tab:upstream} and Table \ref{tab:downstream}. 

\myparagraph{Unseen class labels} 
in Figure 3 in Sec 4.2 are 12 class labels in downstream datasets: 
dock, 
jetski, 
lift, 
jellyfish, 
stingray, 
CoW, 
chanterelle, 
aeroplane, 
diningtable, 
motorbike, 
pottedplant, 
tvmonitor --
which are not in upstream datasets.

\begin{table}[!t]
  \centering
  \resizebox{0.9\linewidth}{!}{%
  \begin{tabular}{lllll}
    \toprule
    Model & Dataset(s) & Iterations & batch size & learning rate\\
    \midrule
    \multirow{4}{*}{baseline} 
    & L & 90k & 64 & 0.0002 \\
    & C & 90k & 64 & 0.0002 \\
    & O365 & 540k & 64 & 0.0002 \\
    & OID & 540k & 64 & 0.0002 \\
    \midrule 
    \multirow{3}{*}{\bf ScaleDet} 
    & L,C & 90k & 64 & 0.0002 \\
    & L,C,O365 & 400k & 128 & 0.0002 \\
    & L,C,O365,OID & 750k & 128 & 0.0002 \\
    \bottomrule
  \end{tabular}
  }
  \vskip -1em
  \caption{Learning schedules on different setups.} 
\label{tab:1}
\vskip -2em
\end{table}

\myparagraph{Training details.}
We provide our pseudo code in Algorithm \ref{alg:code}. 
We will release our pre-trained model weights if our paper is accepted. Our codebase is proprietary, so we could not provide it in supplementary materials. 
Similar to most state-of-the-art detectors, our implementation is built upon Detectron2 \cite{wu2019detectron2}, so it is easy to reproduce our results based upon the codebase. In Eq. (8), we set the hyperparameter $\lambda$ to 10 based on validation set. For the text encoder, we use ``ViT-B/32'' in CLIP \cite{radford2021learning}\footnote{\url{https://github.com/openai/CLIP}} or ``ViT-H/14" in OpenCLIP \cite{ilharco_gabriel_2021_5143773}\footnote{\url{https://github.com/mlfoundations/open_clip}}.

We detail the learning schedules in Sec 4.2 in the following for reproducibility. 
For Table 1, the learning schedules are in Table \ref{tab:1}. 
For Table 2, 4, 5, 6, 7, 8, the learning schedules are the same as the ones trained on the same datasets in Table \ref{tab:1}. For Table 3, we follow the same learning schedule as UniDet \cite{zhou2022simple} to train on three datasets for fair comparison. 

\myparagraph{Complexity analysis.}  The label space of ScaleDet is computed before training and does not require additional computation cost for training. Compared to a standard detector using one loss for classification, ScaleDet uses two loss terms (Eq. (6), Eq. (7)) for classification. Eq. (6) has the same computation cost as standard classification loss, Eq. (7) introduces small computation cost with complexity $O(N)$, proportional to the batch size. At test time, the computation cost of ScaleDet is the same as a standard detector.

\begin{table*}[!t]
  \centering
  \resizebox{\linewidth}{!}{%
  \begin{tabular}{l | l | l | ccccccccccccc c}
    \toprule
    Model & Model Type & \#Data & AerialDrone & Aquarium & Rabbits & EgoHands & Mushrooms & Packages & PascalVOC & pistols & pothole & Raccoon & Shellfish & Thermal & Vehicles & mAP \\
    \midrule
    GLIP-T\cite{li2022grounded} & \multirow{3}{*}{\makecell[l]{detection +\\  understanding}} & 5.5M & 31.2 & 52.5 & 70.8 & 78.7 & 88.1 & 75.6 & 62.3 & 71.2 & 58.7 & 61.4 & 51.4 & 76.7 & 65.3 & 64.9 \\
    GLIPv2-T\cite{zhang2022glipv2} & & 5.5M & 30.2 & 52.5 & 74.8 & 80.0 & 88.1 & 74.3 & 66.4 & 73.0 & 60.1 & 63.7 & 54.4 & 63.0 & 83.5 & 66.5 \\ 
    GLIPv2-B\cite{zhang2022glipv2} & & 20.5M & 32.6 & 57.5 & 73.6 & 80.0 & 88.1 & 74.9 & 71.1 & 76.5 & 58.7 & 68.2 & 70.6 & 79.6 & 71.2 & 69.4 \\
    \midrule
    Detic-R\cite{zhou2022detecting} & \multirow{2}{*}{\makecell[l]{detection +\\  classification}} & 12.6M & 43.1 & 49.5 & 72.2 & 78.0 & 91.0 & 70.0 & 49.1 & 99.9 & 45.8 & 66.2 & 38.0 & 82.3 & 52.4 & 64.4 \\
    Detic-B\cite{zhou2022detecting} & & 12.6M & 44.3 & 51.9 & 76.2 & 78.2 & 89.8 & 75.1 & 96.4 & 99.9 & 50.7 & 64.3 & 47.6 & 82.4 & 54.4 & 70.1 \\
    \midrule
    ScaleDet-R & \multirow{3}{*}{\makecell[l]{detection}} & 3.6M & 42.6 & 49.2 & 72.3 & 78.1 & 89.8 & 75.7 & 91.8 & 99.9 & 47.4 & 67.5 & 40.0 & 82.1 & 54.5 & 68.5 \\
    ScaleDet-T & & 3.6M & 46.0 & 52.1 & 73.1 & 79.2 & 91.6 & 76.5 & 95.1 & 99.9 & 50.1 & 65.6 & 47.8 & 83.1 & 54.7 & 70.4 \\ 
    ScaleDet-B & & 3.6M & 44.8 & 53.5 & 75.1 & 78.4 & 91.0 & 76.5 & 95.7 & 99.9 & 51.7 & 67.6 & 55.0 & 83.4 & 60.7 & \bf 71.8 \\
    \bottomrule
  \end{tabular}
  }
  \vskip -0.5em
  \caption{{\bf Results of fine-tune transfer on 13 individual datasets on ODinW.} 
  R, T, B: ResNet50, Swin-Tiny, Swin-Base Transformer as backbone. 
  Metric: mAP over 13 datasets. Note: ``understanding'', ``classification'' mean training with vision-and-language understanding, and classification datasets besides detection data.
  }
  \label{tab:odinw13}
\vskip -1em
\end{table*}

 \begin{table*}[!t]
   \centering
   \resizebox{0.9\linewidth}{!}{%
   \begin{tabular}{lll}
     \toprule
     Dataset & License & Website \\
     \midrule
     AerialMaritimeDrone  & MIT & \url{https://public.roboflow.com/object-detection/aerial-maritime} \\
     Aquarium & CC BY 4.0 & \url{https://public.roboflow.com/object-detection/aquarium} \\
     CottontailRabbits & CC BY 4.0 &\url{https://public.roboflow.com/object-detection/cottontail-rabbits-video-dataset} \\
     EgoHands & CC BY 4.0 & \url{https://public.roboflow.com/object-detection/hands} \\
     NorthAmericaMushrooms & Public Domain & \url{https://public.roboflow.com/object-detection/na-mushrooms} \\
     Packages & Public Domain & \url{https://public.roboflow.com/object-detection/packages-dataset} \\
     PascalVOC & CC BY 4.0 & \url{https://public.roboflow.com/object-detection/pascal-voc-2012} \\
     pistols & Public Domain & \url{https://public.roboflow.com/object-detection/pistols} \\
     pothole & ODbL v1.0 & \url{https://public.roboflow.com/object-detection/pothole} \\
     Raccoon & MIT & \url{https://public.roboflow.com/object-detection/raccoon} \\
     ShellfishOpenImages & CC BY 4.0 & \url{https://public.roboflow.com/object-detection/shellfish-openimages} \\
     thermalDogsAndPeople & Public Domain & \url{https://public.roboflow.com/object-detection/thermal-dogs-and-people} \\
     VehiclesOpenImages & CC BY 4.0 & \url{https://public.roboflow.com/object-detection/vehicles-openimages} \\
     \bottomrule
   \end{tabular}
   }
   \vskip -0.5em
   \caption{Dataset details on 13 downstream datasets from ``Object Detection in the Wild'' (ODinW-13) benchmark.}
 \label{tab:downstream}
 \end{table*}

\begin{algorithm*}
    \caption{Scalable multi-dataset training recipe for object detection}
    \textbf{Required}: Datasets $D_1, D_2, ..., D_K$ and their concatenated label spaces $L = L_1 \oplus L_2 \oplus {...} \oplus L_K$, which gives $n$ class labels: $ \{l_1, l_2,..., l_n\}$. \\
    \textbf{Required}: Text encoder from pre-trained CLIP \cite{radford2021learning} or OpenCLIP \cite{ilharco_gabriel_2021_5143773}. Compute the text embedding for each label with prompt engineering\footnote{\url{https://github.com/openai/CLIP/blob/main/notebooks/Prompt_Engineering_for_ImageNet.ipynb}} which gives $n$ text embeddings: $ \{t_1, t_2,..., t_n\}$ for the corresponding labels $ \{l_1, l_2,..., l_n\}$. \\
    \textbf{Required}: An object detector $\theta_{\text{detector}}$ for training.
    \begin{algorithmic}[0]
    \For{$i \gets 1$ to $n$}
    	\For{$j \gets 1$ to $n$}
    		\State $\text{cos}(t_i,t_j)=\frac{t_i \cdot t_j}{||t_i||||t_j||}$ \algorithmiccomment{Compute cosine similarity for text embeddings $t_i, t_j$}
	\EndFor
	\State $\alpha_i=\text{min}\{ \text{cos}(t_i,t_j) \}_{j=1}^{n}, \beta_i=\text{max}\{ \text{cos}(t_i,t_j) \}_{j=1}^{n}=\text{cos}(t_i,t_i)=1$ \algorithmiccomment{Compute normalization factors}
	\State $\mathbf{s}_i=\text{sim}(l_i,l_j)=\frac{\text{cos}(t_i,t_j)-\alpha_i}{\beta_i-\alpha_i}$ \algorithmiccomment{Normalize the similarity score within 0 and 1}
    \EndFor
    \end{algorithmic}
    \textbf{Input}: A mini-batch of images $[I_1, I_2, ..., I_B]$, text embeddings $[t_1, t_2, ..., t_B]$ that represent their class labels $[y_1, y_2, ..., y_B]$.
    \begin{algorithmic}[1]
    \For{$b \gets 1$ to $B$}
        \State {$\{v_1, v_2, ..., v_j\} = \theta_{\text{detector}}(I_b)$}   \algorithmiccomment{Obtain visual features from the detector for image $I_b$}
        \For{$i \gets 1$ to $j$}
        		\State {$\mathbf{c}_i = [ \text{cos}(v_i,t_1), \text{cos}(v_i,t_2), ..., \text{cos}(v_i,t_n)]$} \algorithmiccomment{Compute the visual-langague similarities: Eq. (5)} 
		\State {$\mathcal{L}_{hl} = \text{BCE}(\sigma_{sg}(\mathbf{c}_i/\tau), l_i)$} \algorithmiccomment{Compute the hard label assignment loss: Eq. (6)}
		\State {$\mathcal{L}_{sl} = \text{MSE}(\mathbf{c}_i, \mathbf{s}_i)$} \algorithmiccomment{Compute the soft label assignment loss: Eq. (6)}
		\State {$\mathcal{L}_{lang} = \mathcal{L}_{hl} + \lambda \mathcal{L}_{sl}$} \algorithmiccomment{Compute the learning objective on visual feature $v_i$: Eq. (7)}
    \EndFor
     \State {Backward propagation on the detector $\theta_{\text{detector}}(\cdot)$ with the overall loss on all visual features $\{v_1, v_2, ..., v_j\}$}
    \EndFor
    \end{algorithmic}
    \label{alg:code}
\end{algorithm*}

\newpage
{\small
\bibliographystyle{ieee_fullname}
\bibliography{reference}

\begin{thebibliography}{10}\itemsep=-1pt

\bibitem{bansal2018zero}
Ankan Bansal, Karan Sikka, Gaurav Sharma, Rama Chellappa, and Ajay Divakaran.
\newblock Zero-shot object detection.
\newblock In {\em ECCV}, 2018.

\bibitem{caba2015activitynet}
Fabian Caba~Heilbron, Victor Escorcia, Bernard Ghanem, and Juan Carlos~Niebles.
\newblock Activitynet: A large-scale video benchmark for human activity
  understanding.
\newblock In {\em CVPR}, 2015.

\bibitem{cai2022x}
Zhaowei Cai, Gukyeong Kwon, Avinash Ravichandran, Erhan Bas, Zhuowen Tu, Rahul
  Bhotika, and Stefano Soatto.
\newblock X-detr: A versatile architecture for instance-wise vision-language
  tasks.
\newblock In {\em ECCV}, 2022.

\bibitem{cai2018cascade}
Zhaowei Cai and Nuno Vasconcelos.
\newblock Cascade r-cnn: Delving into high quality object detection.
\newblock In {\em CVPR}, 2018.

\bibitem{carion2020end}
Nicolas Carion, Francisco Massa, Gabriel Synnaeve, Nicolas Usunier, Alexander
  Kirillov, and Sergey Zagoruyko.
\newblock End-to-end object detection with transformers.
\newblock In {\em ECCV}, 2020.

\bibitem{carreira2017quo}
Joao Carreira and Andrew Zisserman.
\newblock Quo vadis, action recognition? a new model and the kinetics dataset.
\newblock In {\em CVPR}, 2017.

\bibitem{changpinyo2021conceptual}
Soravit Changpinyo, Piyush Sharma, Nan Ding, and Radu Soricut.
\newblock Conceptual 12m: Pushing web-scale image-text pre-training to
  recognize long-tail visual concepts.
\newblock In {\em CVPR}, 2021.

\bibitem{dai2021dynamic}
Xiyang Dai, Yinpeng Chen, Bin Xiao, Dongdong Chen, Mengchen Liu, Lu Yuan, and
  Lei Zhang.
\newblock Dynamic head: Unifying object detection heads with attentions.
\newblock In {\em CVPR}, 2021.

\bibitem{deng2009imagenet}
Jia Deng, Wei Dong, Richard Socher, Li-Jia Li, Kai Li, and Li Fei-Fei.
\newblock Imagenet: A large-scale hierarchical image database.
\newblock In {\em CVPR}, 2009.

\bibitem{desai2021virtex}
Karan Desai and Justin Johnson.
\newblock Virtex: Learning visual representations from textual annotations.
\newblock In {\em CVPR}, 2021.

\bibitem{ding2022open}
Zheng Ding, Jieke Wang, and Zhuowen Tu.
\newblock Open-vocabulary panoptic segmentation with maskclip.
\newblock {\em arXiv preprint arXiv:2208.08984}, 2022.

\bibitem{ghiasi2021simple}
Golnaz Ghiasi, Yin Cui, Aravind Srinivas, Rui Qian, Tsung-Yi Lin, Ekin~D Cubuk,
  Quoc~V Le, and Barret Zoph.
\newblock Simple copy-paste is a strong data augmentation method for instance
  segmentation.
\newblock In {\em CVPR}, 2021.

\bibitem{gu2021open}
Xiuye Gu, Tsung-Yi Lin, Weicheng Kuo, and Yin Cui.
\newblock Open-vocabulary object detection via vision and language knowledge
  distillation.
\newblock In {\em ICLR}, 2021.

\bibitem{gupta2019lvis}
Agrim Gupta, Piotr Dollar, and Ross Girshick.
\newblock Lvis: A dataset for large vocabulary instance segmentation.
\newblock In {\em CVPR}, 2019.

\bibitem{he2017mask}
Kaiming He, Georgia Gkioxari, Piotr Doll{\'a}r, and Ross Girshick.
\newblock Mask r-cnn.
\newblock In {\em ICCV}, 2017.

\bibitem{he2016deep}
Kaiming He, Xiangyu Zhang, Shaoqing Ren, and Jian Sun.
\newblock Deep residual learning for image recognition.
\newblock In {\em CVPR}, 2016.

\bibitem{hu2021unit}
Ronghang Hu and Amanpreet Singh.
\newblock Unit: Multimodal multitask learning with a unified transformer.
\newblock In {\em ICCV}, 2021.

\bibitem{ilharco_gabriel_2021_5143773}
Gabriel Ilharco, Mitchell Wortsman, Ross Wightman, Cade Gordon, Nicholas
  Carlini, Rohan Taori, Achal Dave, Vaishaal Shankar, Hongseok Namkoong, John
  Miller, Hannaneh Hajishirzi, Ali Farhadi, and Ludwig Schmidt.
\newblock Openclip.
\newblock {\em https://doi.org/10.5281/zenodo.5143773}, 2021.

\bibitem{jia2021scaling}
Chao Jia, Yinfei Yang, Ye Xia, Yi-Ting Chen, Zarana Parekh, Hieu Pham, Quoc Le,
  Yun-Hsuan Sung, Zhen Li, and Tom Duerig.
\newblock Scaling up visual and vision-language representation learning with
  noisy text supervision.
\newblock In {\em ICML}, 2021.

\bibitem{kamath2021mdetr}
Aishwarya Kamath, Mannat Singh, Yann LeCun, Gabriel Synnaeve, Ishan Misra, and
  Nicolas Carion.
\newblock Mdetr-modulated detection for end-to-end multi-modal understanding.
\newblock In {\em ICCV}, 2021.

\bibitem{kingma2014adam}
Diederik~P Kingma and Jimmy Ba.
\newblock Adam: A method for stochastic optimization.
\newblock {\em arXiv preprint arXiv:1412.6980}, 2014.

\bibitem{OpenImages2}
Ivan Krasin, Tom Duerig, Neil Alldrin, Vittorio Ferrari, Sami Abu-El-Haija,
  Alina Kuznetsova, Hassan Rom, Jasper Uijlings, Stefan Popov, Shahab Kamali,
  Matteo Malloci, Jordi Pont-Tuset, Andreas Veit, Serge Belongie, Victor Gomes,
  Abhinav Gupta, Chen Sun, Gal Chechik, David Cai, Zheyun Feng, Dhyanesh
  Narayanan, and Kevin Murphy.
\newblock Openimages: A public dataset for large-scale multi-label and
  multi-class image classification.
\newblock {\em Dataset available from
  https://storage.googleapis.com/openimages/web/index.html}, 2017.

\bibitem{krishna2017visual}
Ranjay Krishna, Yuke Zhu, Oliver Groth, Justin Johnson, Kenji Hata, Joshua
  Kravitz, Stephanie Chen, Yannis Kalantidis, Li-Jia Li, David~A Shamma, et~al.
\newblock Visual genome: Connecting language and vision using crowdsourced
  dense image annotations.
\newblock {\em IJCV}, 2017.

\bibitem{kuznetsova2020open}
Alina Kuznetsova, Hassan Rom, Neil Alldrin, Jasper Uijlings, Ivan Krasin, Jordi
  Pont-Tuset, Shahab Kamali, Stefan Popov, Matteo Malloci, Alexander
  Kolesnikov, et~al.
\newblock The open images dataset v4.
\newblock {\em IJCV}, 2020.

\bibitem{lambert2020mseg}
John Lambert, Zhuang Liu, Ozan Sener, James Hays, and Vladlen Koltun.
\newblock Mseg: A composite dataset for multi-domain semantic segmentation.
\newblock In {\em CVPR}, 2020.

\bibitem{li2022grounded}
Liunian~Harold Li, Pengchuan Zhang, Haotian Zhang, Jianwei Yang, Chunyuan Li,
  Yiwu Zhong, Lijuan Wang, Lu Yuan, Lei Zhang, Jenq-Neng Hwang, et~al.
\newblock Grounded language-image pre-training.
\newblock In {\em CVPR}, 2022.

\bibitem{lin2014microsoft}
Tsung-Yi Lin, Michael Maire, Serge Belongie, James Hays, Pietro Perona, Deva
  Ramanan, Piotr Doll{\'a}r, and C~Lawrence Zitnick.
\newblock Microsoft coco: Common objects in context.
\newblock In {\em ECCV}, 2014.

\bibitem{liu2021swin}
Ze Liu, Yutong Lin, Yue Cao, Han Hu, Yixuan Wei, Zheng Zhang, Stephen Lin, and
  Baining Guo.
\newblock Swin transformer: Hierarchical vision transformer using shifted
  windows.
\newblock In {\em ICCV}, 2021.

\bibitem{ordonez2011im2text}
Vicente Ordonez, Girish Kulkarni, and Tamara Berg.
\newblock Im2text: Describing images using 1 million captioned photographs.
\newblock In {\em NeurIPS}, 2011.

\bibitem{plummer2015flickr30k}
Bryan~A Plummer, Liwei Wang, Chris~M Cervantes, Juan~C Caicedo, Julia
  Hockenmaier, and Svetlana Lazebnik.
\newblock Flickr30k entities: Collecting region-to-phrase correspondences for
  richer image-to-sentence models.
\newblock In {\em ICCV}, 2015.

\bibitem{radford2021learning}
Alec Radford, Jong~Wook Kim, Chris Hallacy, Aditya Ramesh, Gabriel Goh,
  Sandhini Agarwal, Girish Sastry, Amanda Askell, Pamela Mishkin, Jack Clark,
  et~al.
\newblock Learning transferable visual models from natural language
  supervision.
\newblock In {\em ICML}, 2021.

\bibitem{ren2015faster}
Shaoqing Ren, Kaiming He, Ross Girshick, and Jian Sun.
\newblock Faster r-cnn: Towards real-time object detection with region proposal
  networks.
\newblock In {\em NeurIPS}, 2015.

\bibitem{ridnik2021imagenet}
Tal Ridnik, Emanuel Ben-Baruch, Asaf Noy, and Lihi Zelnik-Manor.
\newblock Imagenet-21k pretraining for the masses.
\newblock {\em arXiv preprint arXiv:2104.10972}, 2021.

\bibitem{shao2019objects365}
Shuai Shao, Zeming Li, Tianyuan Zhang, Chao Peng, Gang Yu, Xiangyu Zhang, Jing
  Li, and Jian Sun.
\newblock Objects365: A large-scale, high-quality dataset for object detection.
\newblock In {\em ICCV}, 2019.

\bibitem{sharma2018conceptual}
Piyush Sharma, Nan Ding, Sebastian Goodman, and Radu Soricut.
\newblock Conceptual captions: A cleaned, hypernymed, image alt-text dataset
  for automatic image captioning.
\newblock In {\em ACL}, 2018.

\bibitem{tian2019fcos}
Zhi Tian, Chunhua Shen, Hao Chen, and Tong He.
\newblock Fcos: Fully convolutional one-stage object detection.
\newblock In {\em ICCV}, 2019.

\bibitem{uijlings2022missing}
Jasper Uijlings, Thomas Mensink, and Vittorio Ferrari.
\newblock The missing link: Finding label relations across datasets.
\newblock In {\em ECCV}, 2022.

\bibitem{wang2019towards}
Xudong Wang, Zhaowei Cai, Dashan Gao, and Nuno Vasconcelos.
\newblock Towards universal object detection by domain attention.
\newblock In {\em CVPR}, 2019.

\bibitem{wu2019detectron2}
Yuxin Wu, Alexander Kirillov, Francisco Massa, Wan-Yen Lo, and Ross Girshick.
\newblock Detectron2.
\newblock \url{https://github.com/facebookresearch/detectron2}, 2019.

\bibitem{xu2020universal}
Hang Xu, Linpu Fang, Xiaodan Liang, Wenxiong Kang, and Zhenguo Li.
\newblock Universal-rcnn: Universal object detector via transferable graph
  r-cnn.
\newblock In {\em AAAI}, 2020.

\bibitem{xu2022groupvit}
Jiarui Xu, Shalini De~Mello, Sifei Liu, Wonmin Byeon, Thomas Breuel, Jan Kautz,
  and Xiaolong Wang.
\newblock Groupvit: Semantic segmentation emerges from text supervision.
\newblock In {\em CVPR}, 2022.

\bibitem{yang2022unified}
Jianwei Yang, Chunyuan Li, Pengchuan Zhang, Bin Xiao, Ce Liu, Lu Yuan, and
  Jianfeng Gao.
\newblock Unified contrastive learning in image-text-label space.
\newblock In {\em CVPR}, 2022.

\bibitem{yao2021filip}
Lewei Yao, Runhui Huang, Lu Hou, Guansong Lu, Minzhe Niu, Hang Xu, Xiaodan
  Liang, Zhenguo Li, Xin Jiang, and Chunjing Xu.
\newblock Filip: Fine-grained interactive language-image pre-training.
\newblock In {\em ICLR}, 2021.

\bibitem{zareian2021open}
Alireza Zareian, Kevin~Dela Rosa, Derek~Hao Hu, and Shih-Fu Chang.
\newblock Open-vocabulary object detection using captions.
\newblock In {\em CVPR}, 2021.

\bibitem{zhai2022lit}
Xiaohua Zhai, Xiao Wang, Basil Mustafa, Andreas Steiner, Daniel Keysers,
  Alexander Kolesnikov, and Lucas Beyer.
\newblock Lit: Zero-shot transfer with locked-image text tuning.
\newblock In {\em CVPR}, 2022.

\bibitem{zhang2022glipv2}
Haotian* Zhang, Pengchuan* Zhang, Xiaowei Hu, Yen-Chun Chen, Liunian~Harold Li,
  Xiyang Dai, Lijuan Wang, Lu Yuan, Jenq-Neng Hwang, and Jianfeng Gao.
\newblock Glipv2: Unifying localization and vision-language understanding.
\newblock In {\em NeurIPS}, 2022.

\bibitem{zhao2020object}
Xiangyun Zhao, Samuel Schulter, Gaurav Sharma, Yi-Hsuan Tsai, Manmohan
  Chandraker, and Ying Wu.
\newblock Object detection with a unified label space from multiple datasets.
\newblock In {\em ECCV}, 2020.

\bibitem{zhong2022regionclip}
Yiwu Zhong, Jianwei Yang, Pengchuan Zhang, Chunyuan Li, Noel Codella,
  Liunian~Harold Li, Luowei Zhou, Xiyang Dai, Lu Yuan, Yin Li, et~al.
\newblock Regionclip: Region-based language-image pretraining.
\newblock In {\em CVPR}, 2022.

\bibitem{zhou2022detecting}
Xingyi Zhou, Rohit Girdhar, Armand Joulin, Phillip Kr{\"a}henb{\"u}hl, and
  Ishan Misra.
\newblock Detecting twenty-thousand classes using image-level supervision.
\newblock In {\em ECCV}, 2022.

\bibitem{zhou2021probabilistic}
Xingyi Zhou, Vladlen Koltun, and Philipp Kr{\"a}henb{\"u}hl.
\newblock Probabilistic two-stage detection.
\newblock {\em arXiv preprint arXiv:2103.07461}, 2021.

\bibitem{zhou2022simple}
Xingyi Zhou, Vladlen Koltun, and Philipp Kr{\"a}henb{\"u}hl.
\newblock Simple multi-dataset detection.
\newblock In {\em CVPR}, 2022.

\bibitem{zhou2019objects}
Xingyi Zhou, Dequan Wang, and Philipp Kr{\"a}henb{\"u}hl.
\newblock Objects as points.
\newblock {\em arXiv preprint arXiv:1904.07850}, 2019.

\end{thebibliography}
}

\end{document}